\documentclass[sigconf]{acmart}
\usepackage{indentfirst}
\usepackage{bm}
\usepackage{multirow}
\usepackage{indentfirst}
\usepackage{algorithm}
\usepackage{algorithmic}

\AtBeginDocument{%
  }

\setcopyright{acmcopyright}
\copyrightyear{2023}
\acmYear{2023}
\acmDOI{XXXXXXX.XXXXXXX}

\acmPrice{15.00}
\acmISBN{978-1-4503-XXXX-X/18/06}

\renewcommand\footnotetextcopyrightpermission[1]{}
\begin{document}
\setlength{\parindent}{2em}

\title{Inter-frame Accelerate Attack against Video Interpolation Models}

\author{Junpei Liao}
\authornote{Both authors contributed equally to this research.}
\authornote{This work is done during Liao's intern at SUN YAT-SEN UNIVERSITY}
\email{junpeiliao@163.com}
\affiliation{%
  \institution{SUN YAT-SEN UNIVERSITY}
  \city{Shenzhen}
  \state{Guangdong}
  \country{China}
}

\author{Zhikai Chen}
\email{zhikai_chen@outlook.com}
\authornotemark[1]
\affiliation{%
  \institution{Tecent}
  \city{Shenzhen}
  \state{Guangdong}
  \country{China}
}

\author{Liang Yi}
\email{isyiliang@std.uestc.edu.cn}
\affiliation{%
  \institution{University of Electronic Science and Technology of China}
  \city{Chengdu}
  \state{Sichuan}
  \country{China}
}

\author{Wen Yuanyang}
\authornote{Corresponding author.}
\email{yangwy56@mail.sysu.edu.cn}
\affiliation{%
  \institution{SUN YAT-SEN UNIVERSITY}
  \city{Shenzhen}
  \state{Guangdong}
  \country{China}
}

\author{Baoyuan Wu}
\email{wubaoyuan@cuhk.edu.cn}
\affiliation{%
  \institution{The Chinese University of Hong Kong, Shenzhen}
  \city{Shenzhen}
  \state{Guangdong}
  \country{China}
}

\author{Xiaochun Cao}
\email{caoxiaochun@mail.sysu.edu.cn}
\affiliation{%
  \institution{SUN YAT-SEN UNIVERSITY}
  \city{Shenzhen}
  \state{Guangdong}
  \country{China}
}

\renewcommand{\shortauthors}{Liao et al.}

\begin{abstract}
  Deep learning based video frame interpolation (VIF) method, aiming to synthesis the intermediate frames to enhance video quality, have been highly developed in the past few years. This paper investigates the adversarial robustness of VIF models. We apply adversarial attacks to VIF models and find that the VIF models are very vulnerable to adversarial examples. To improve attack efficiency, we suggest to make full use of the property of video frame interpolation task. The intuition is that the gap between adjacent frames would be small, leading to the corresponding adversarial perturbations being similar as well. Then we propose a novel attack method named Inter-frame Accelerate Attack (IAA) that initializes the perturbation as the perturbation for the previous adjacent frame and reduces the number of attack iterations. It is shown that our method can improve attack efficiency greatly while achieving comparable attack performance with traditional methods. Besides, we also extend our method to video recognition models which are higher level vision tasks and achieves great attack efficiency.
\end{abstract}




\maketitle

\section{Introduction}
Deep Neural Networks (DNNs) have been shown the vulnerability against the adversarial examples, which are added imperceptible small perturbations. Recent years, the adversarial robustness of higher level and some lower level vision scenarios such as image classification\cite{Goodfellow, Madry}, semantic segmentation and object detection\cite{xie2017adversarial} and super-resolution\cite{choi2019evaluating} has been investigated.  \\
Meanwhile, video frame interpolation (VFI), a lower level vision task, has emerged as a popular research field in recent years, aiming to achieve video temporal super-resolution by generating smooth transitions between consecutive frames. At the beginning, most algorithms concentrate on motion estimation and motion-compensate frame interpolation, such as \cite{ha2004motion, choi2007motion, kang2007motion}. And the quality of motion estimation determines the performance of video frame interpolation results. With the development of deep learning, various deep learning-based VIF methods emerges\cite{DAIN, CAIN, AdaCoF, CDFI, XVFI, RRIN}. Although there are many VIF algorithms based on deep learning, their robustness to adversarial attack has not been investigated. 

\begin{figure}[t]
	\begin{center}
		\centering
		\includegraphics[width=1.0\linewidth]{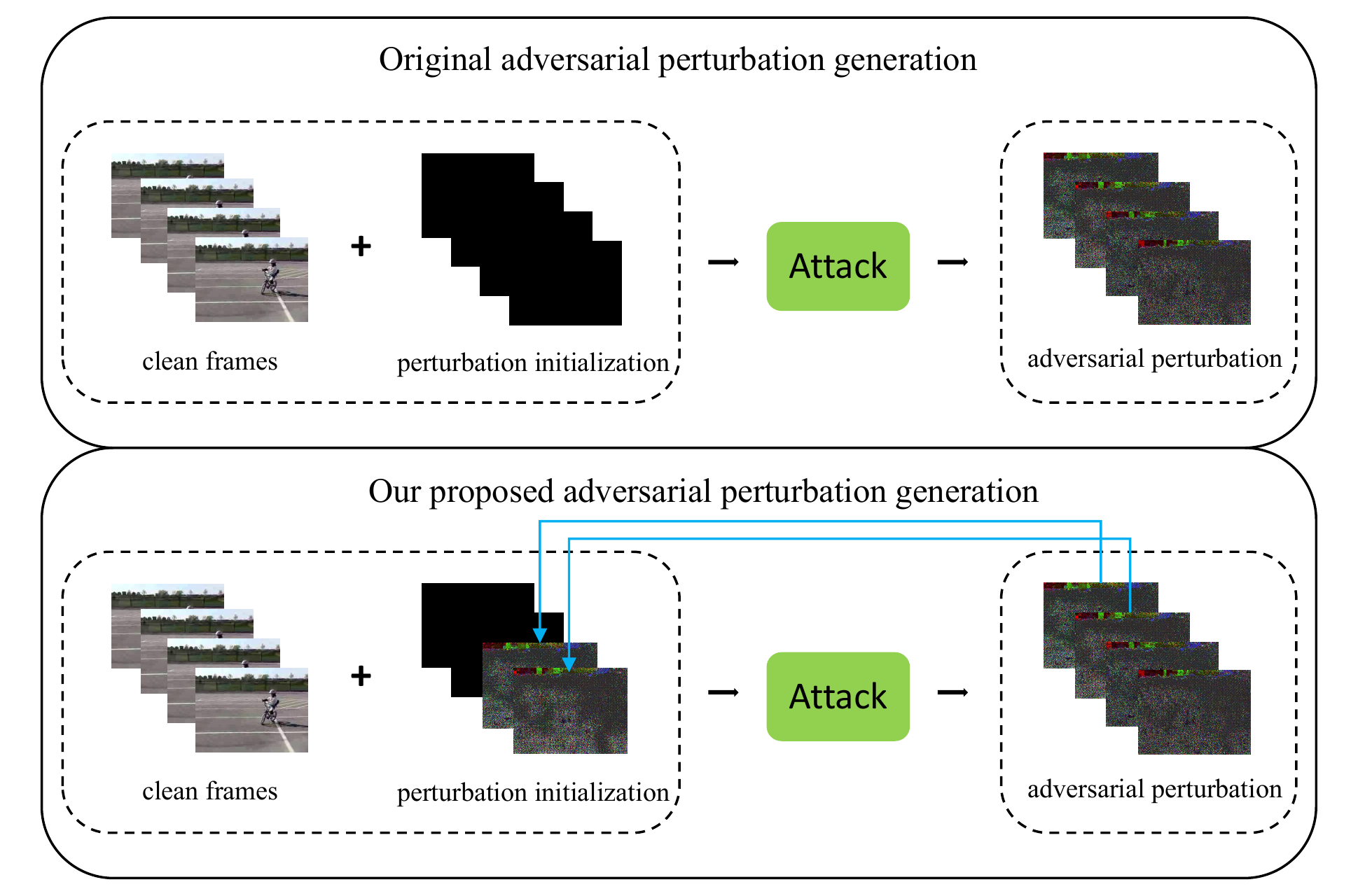}
	\end{center}
	\caption{We demonstrate the differences between original adversarial attack and our proposed method. We initialize the perturbation values of the latter frames as the perturbations generated for the previous ones so that we can inherit the gradient information from the previous similar frames. While taking advantage of the previous generated perturbations, we are able to reduce the attack iteration to accelerate the attack process.}
	\label{fig:pipeline}
\end{figure}

As mentioned above, while most researches focus on image classification and processing models, several video attacks\cite{Wei, chen2021appending, chen2022attacking} are also proposed recently. However, existing video attack methods cannot be applied directly to VIF models since attacking VIF models requires attacker to destruct most of the synthesized intermediate frames in the videos. In this paper, we first evaluate the adversarial robustness of VIF models. We propose a PGD-based attack for VIF models. We generate invisible adversarial perturbations for the previous and next one frame of the intermediate frame which can lead to a great degradation in the quality of synthesized intermediate frames. 

\begin{figure*}[t]
	\begin{center}
		\centering
		\begin{minipage}[b]{0.13\linewidth}
			\centering
			\centerline{\includegraphics[width=1.0\linewidth]{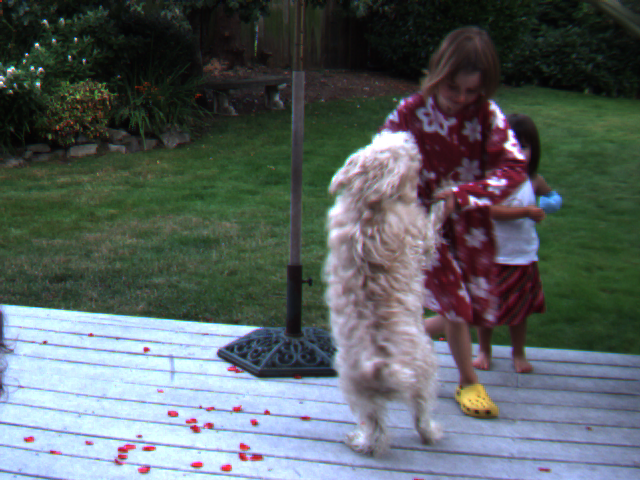}}
		\end{minipage}
  		\begin{minipage}[b]{0.13\linewidth}
			\centering
			\centerline{\includegraphics[width=1.0\linewidth]{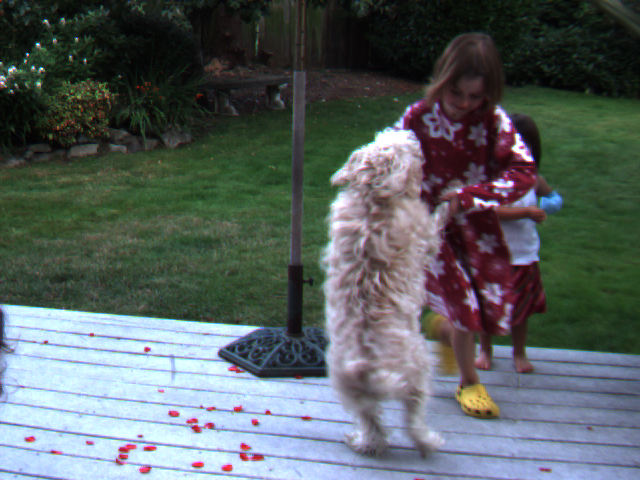}}
		\end{minipage}
		\begin{minipage}[b]{0.13\linewidth}
			\centering
			\centerline{\includegraphics[width=1.0\linewidth]{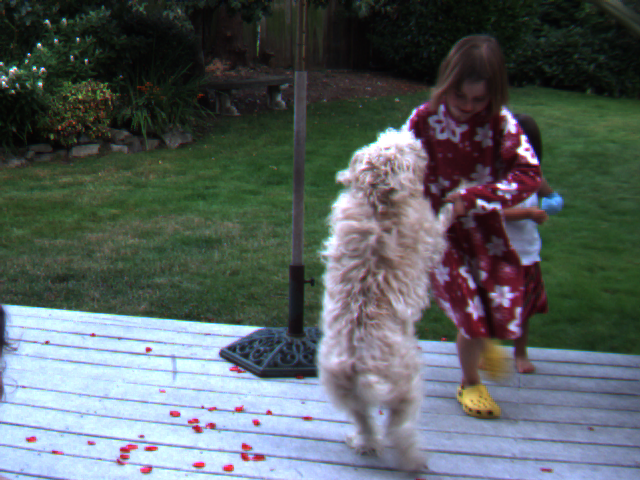}}
		\end{minipage}
  		\begin{minipage}[b]{0.13\linewidth}
			\centering
			\centerline{\includegraphics[width=1.0\linewidth]{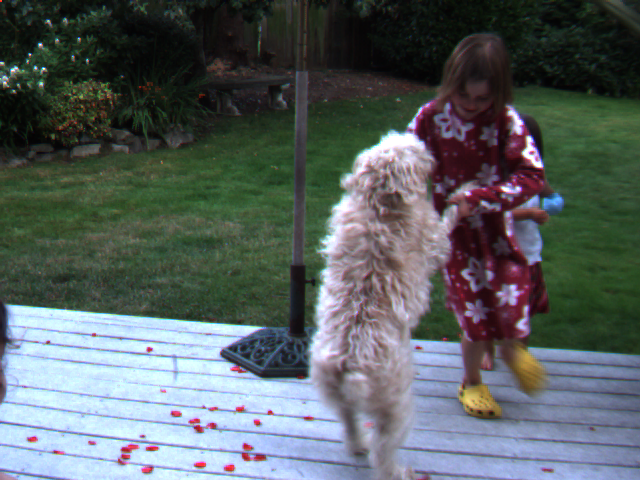}}
		\end{minipage}
  		\begin{minipage}[b]{0.13\linewidth}
			\centering
			\centerline{\includegraphics[width=1.0\linewidth]{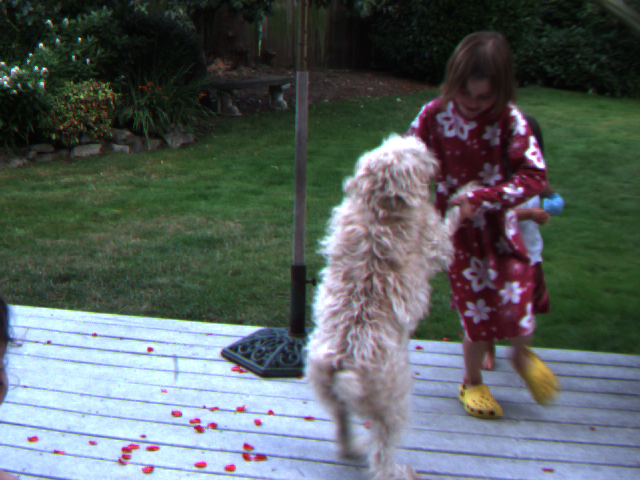}}
		\end{minipage}
  		\begin{minipage}[b]{0.13\linewidth}
			\centering
			\centerline{\includegraphics[width=1.0\linewidth]{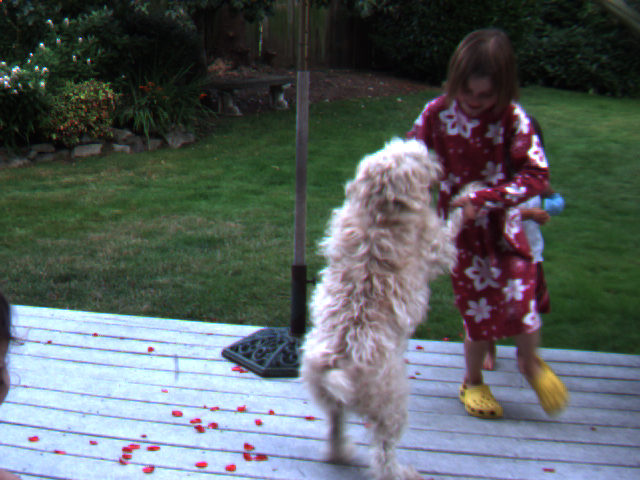}}
		\end{minipage}
		\begin{minipage}[b]{0.13\linewidth}
			\centering
			\centerline{\includegraphics[width=1.0\linewidth]{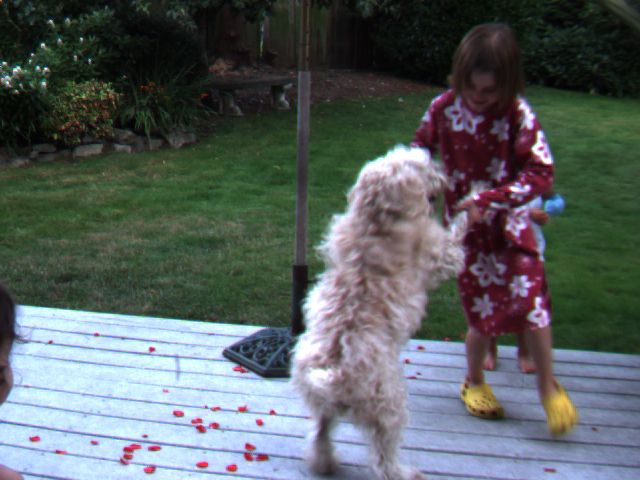}}
		\end{minipage}
  		\begin{minipage}[b]{0.13\linewidth}
			\centering
			\centerline{\includegraphics[width=1.0\linewidth]{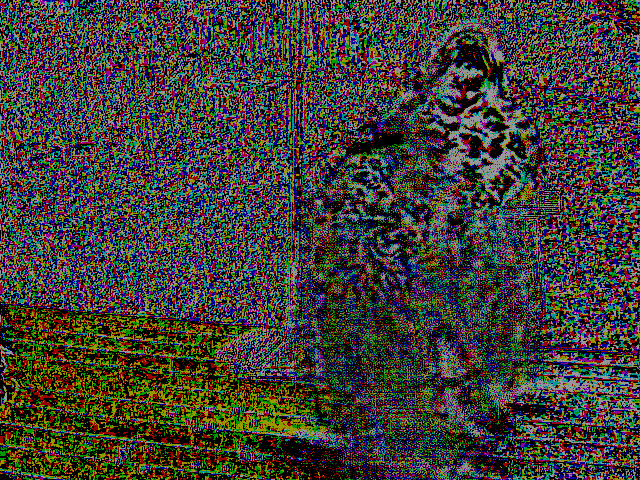}}
		\end{minipage}
    		\begin{minipage}[b]{0.13\linewidth}
			\centering
			\centerline{\includegraphics[width=1.0\linewidth]{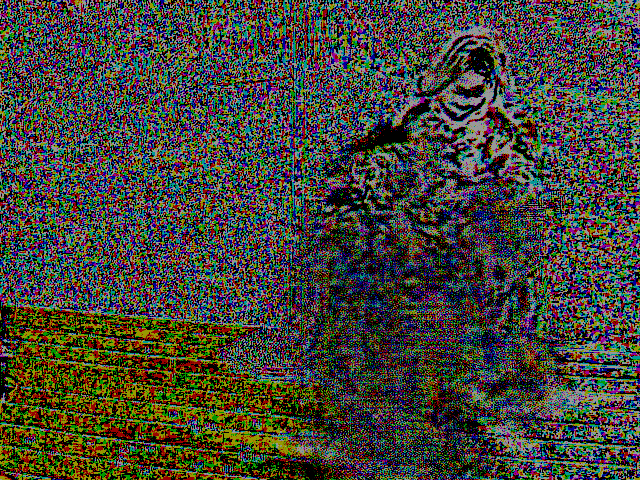}}
		\end{minipage}
		\begin{minipage}[b]{0.13\linewidth}
			\centering
			\centerline{\includegraphics[width=1.0\linewidth]{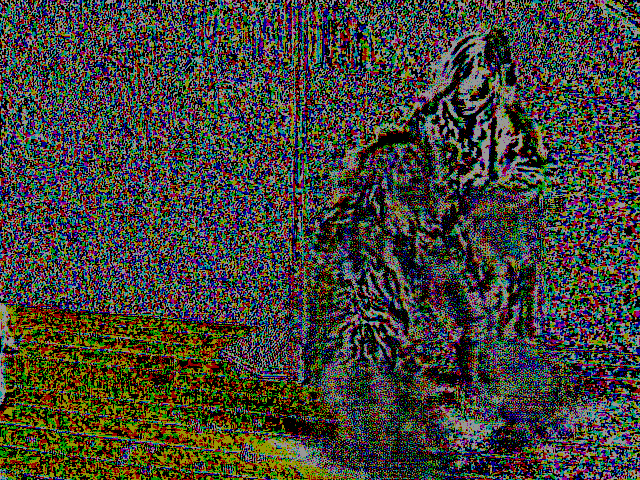}}
		\end{minipage}
    		\begin{minipage}[b]{0.13\linewidth}
			\centering
			\centerline{\includegraphics[width=1.0\linewidth]{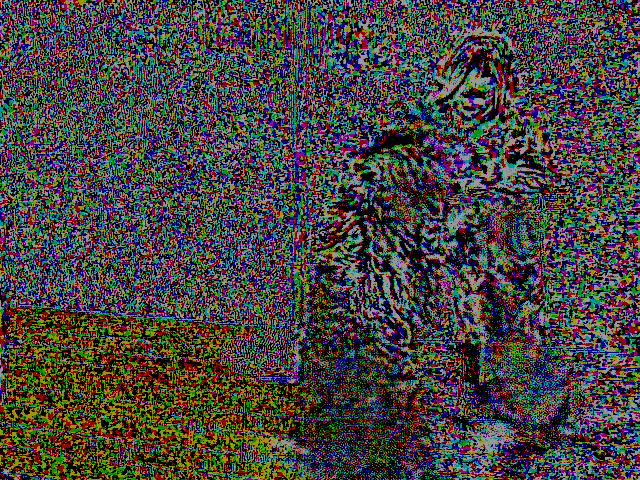}}
		\end{minipage}
  		\begin{minipage}[b]{0.13\linewidth}
			\centering
			\centerline{\includegraphics[width=1.0\linewidth]{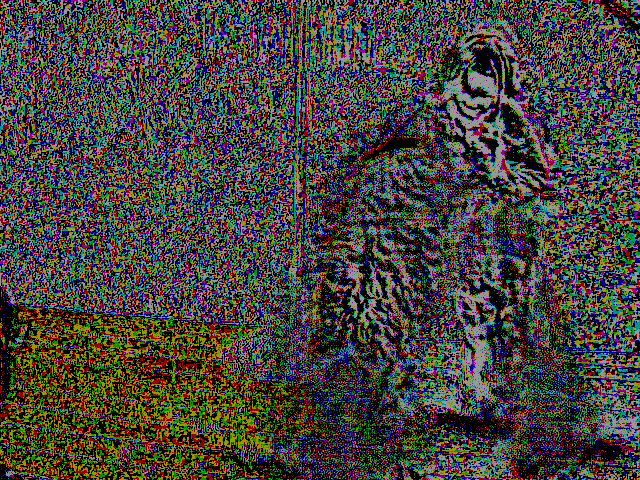}}
		\end{minipage}
    	\begin{minipage}[b]{0.13\linewidth}
			\centering
			\centerline{\includegraphics[width=1.0\linewidth]{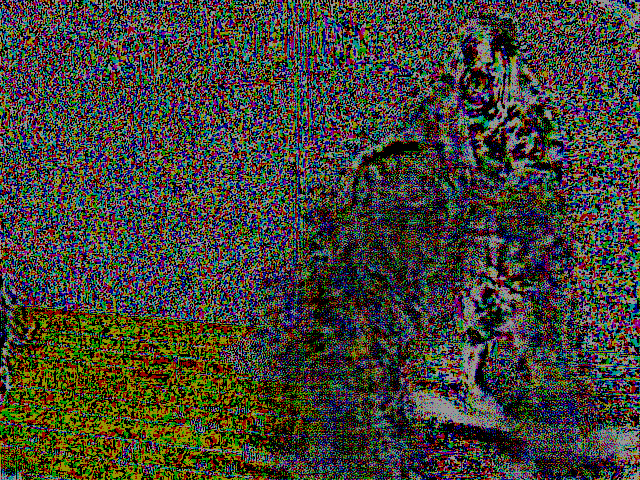}}
		\end{minipage}
		\begin{minipage}[b]{0.13\linewidth}
			\centering
			\centerline{\includegraphics[width=1.0\linewidth]{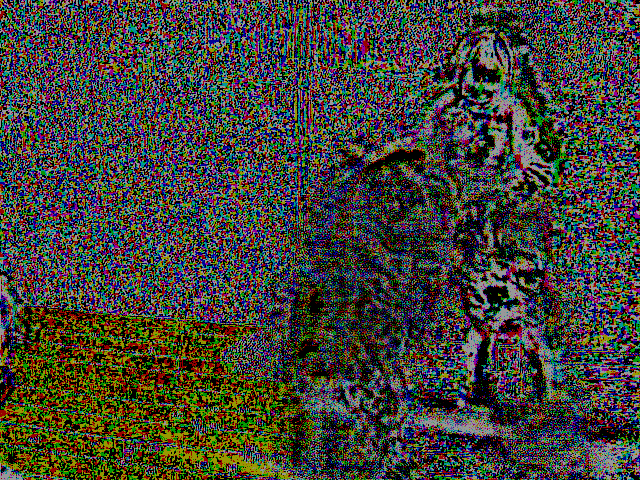}}
		\end{minipage}
	\end{center}
	\caption{Visualization of consecutive frames in the video and corresponding adversarial perturbations generated by basic attack.}
	\label{fig:visual_motivation}
\end{figure*}

\begin{figure*}[t]
	\begin{center}
		\centering
		\begin{minipage}[b]{0.19\linewidth}
			\centering
			\centerline{\includegraphics[width=1.0\linewidth]{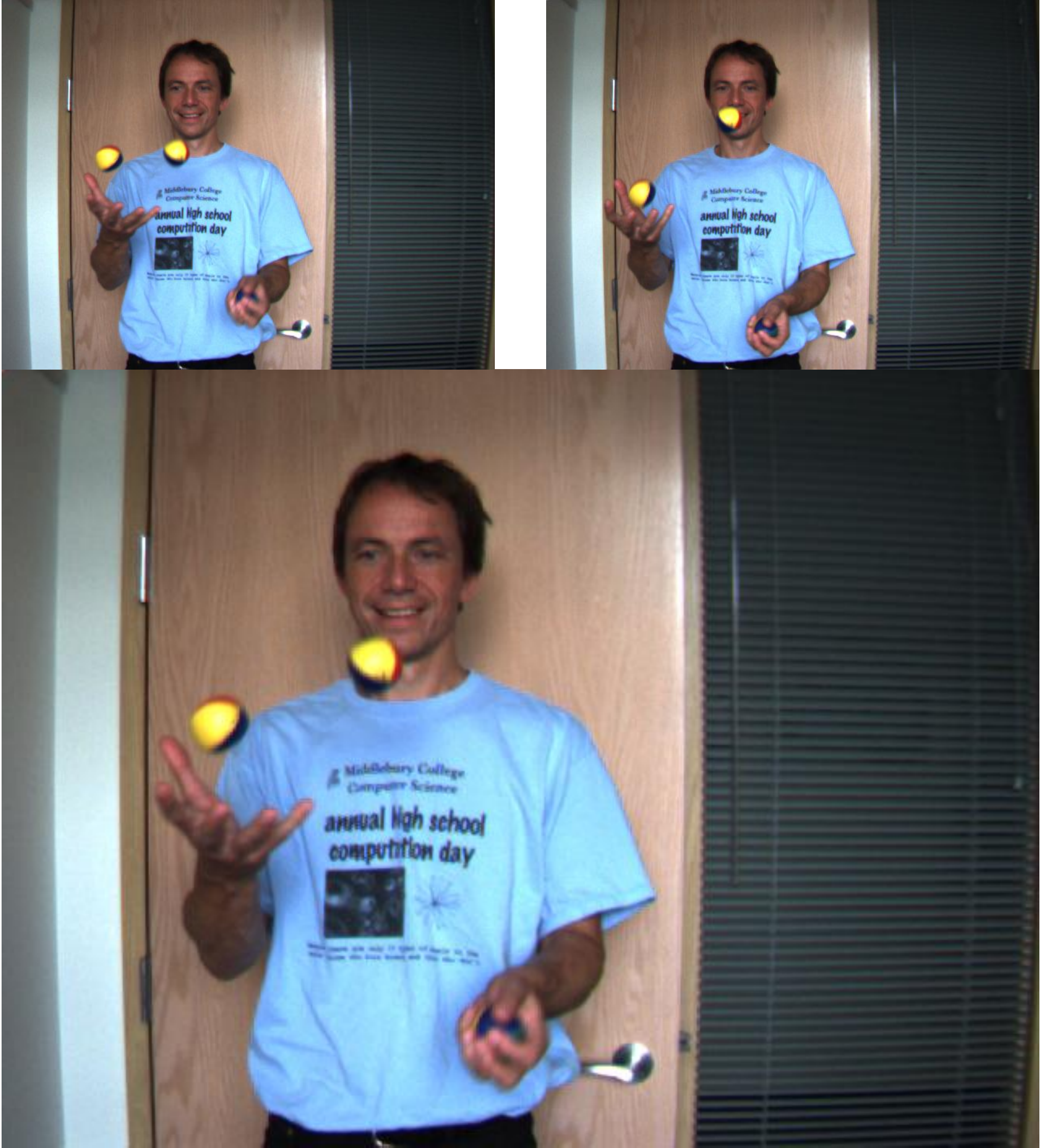}}
			\centerline{Ground truth}
		\end{minipage}
		\begin{minipage}[b]{0.19\linewidth}
			\centering
			\centerline{\includegraphics[width=1.0\linewidth]{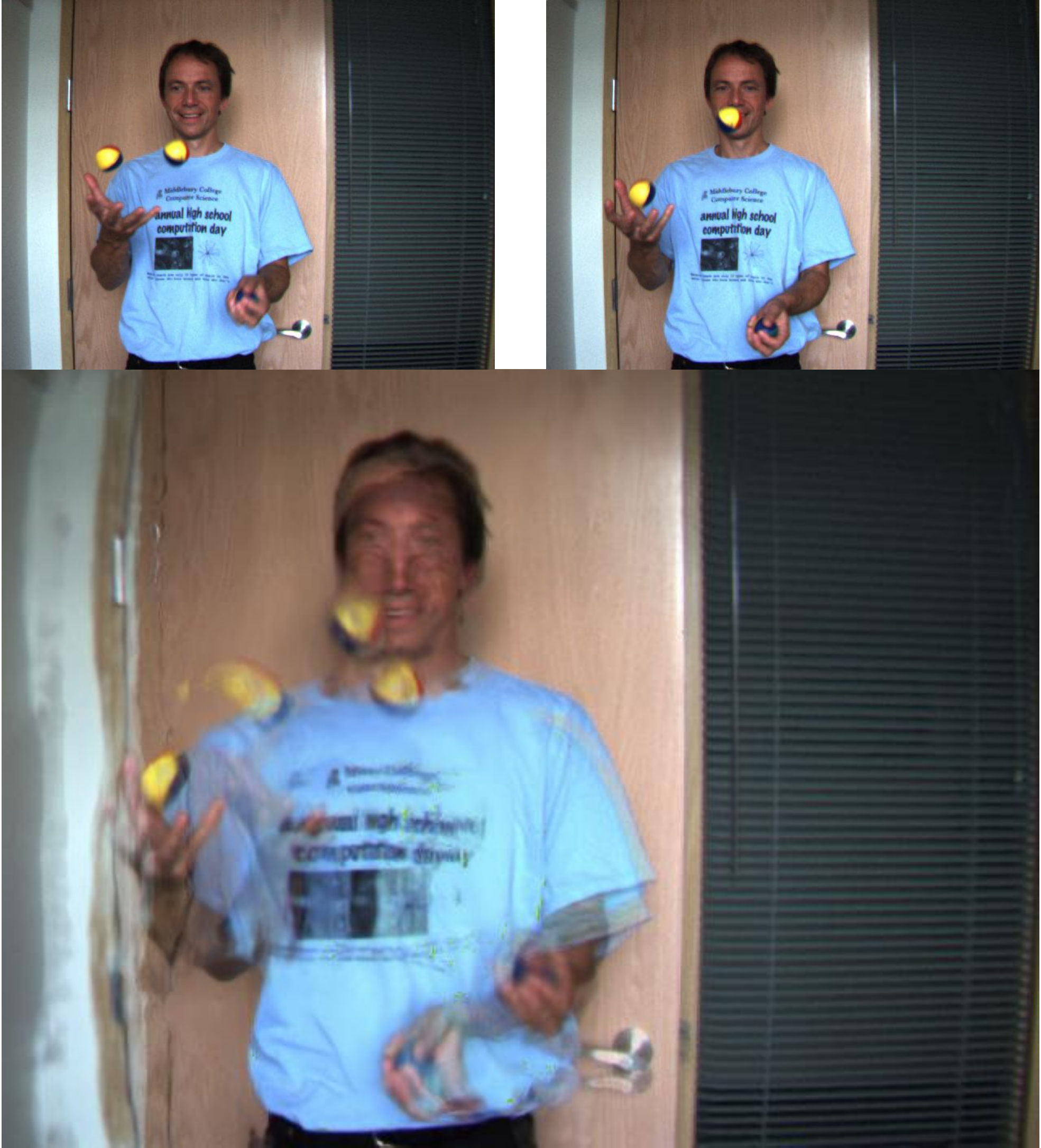}}
			\centerline{$\alpha=0.01$}
		\end{minipage}
		\begin{minipage}[b]{0.19\linewidth}
			\centering
			\centerline{\includegraphics[width=1.0\linewidth]{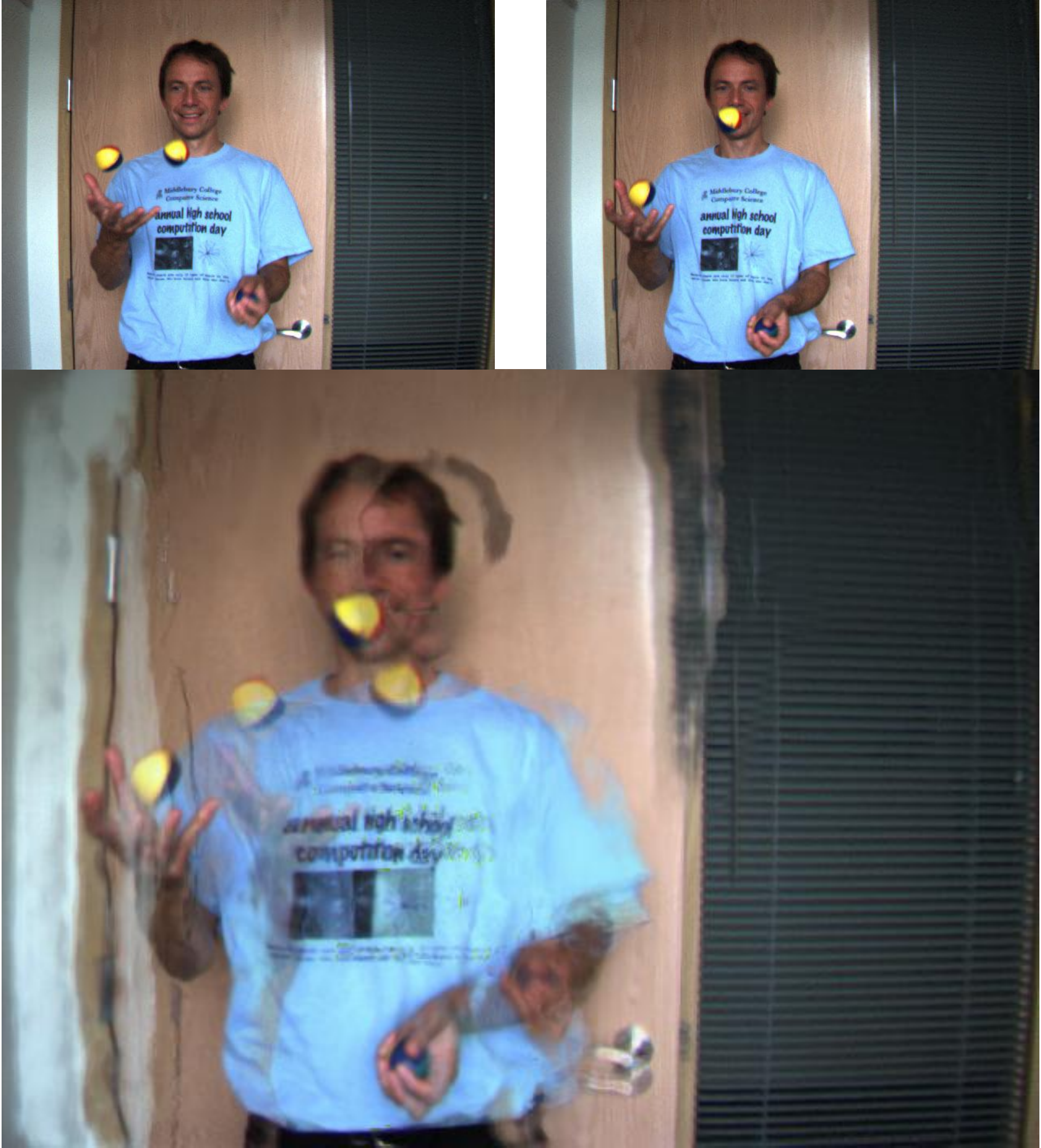}}
			\centerline{$\alpha=0.02$}
		\end{minipage}
		\begin{minipage}[b]{0.19\linewidth}
			\centering
			\centerline{\includegraphics[width=1.0\linewidth]{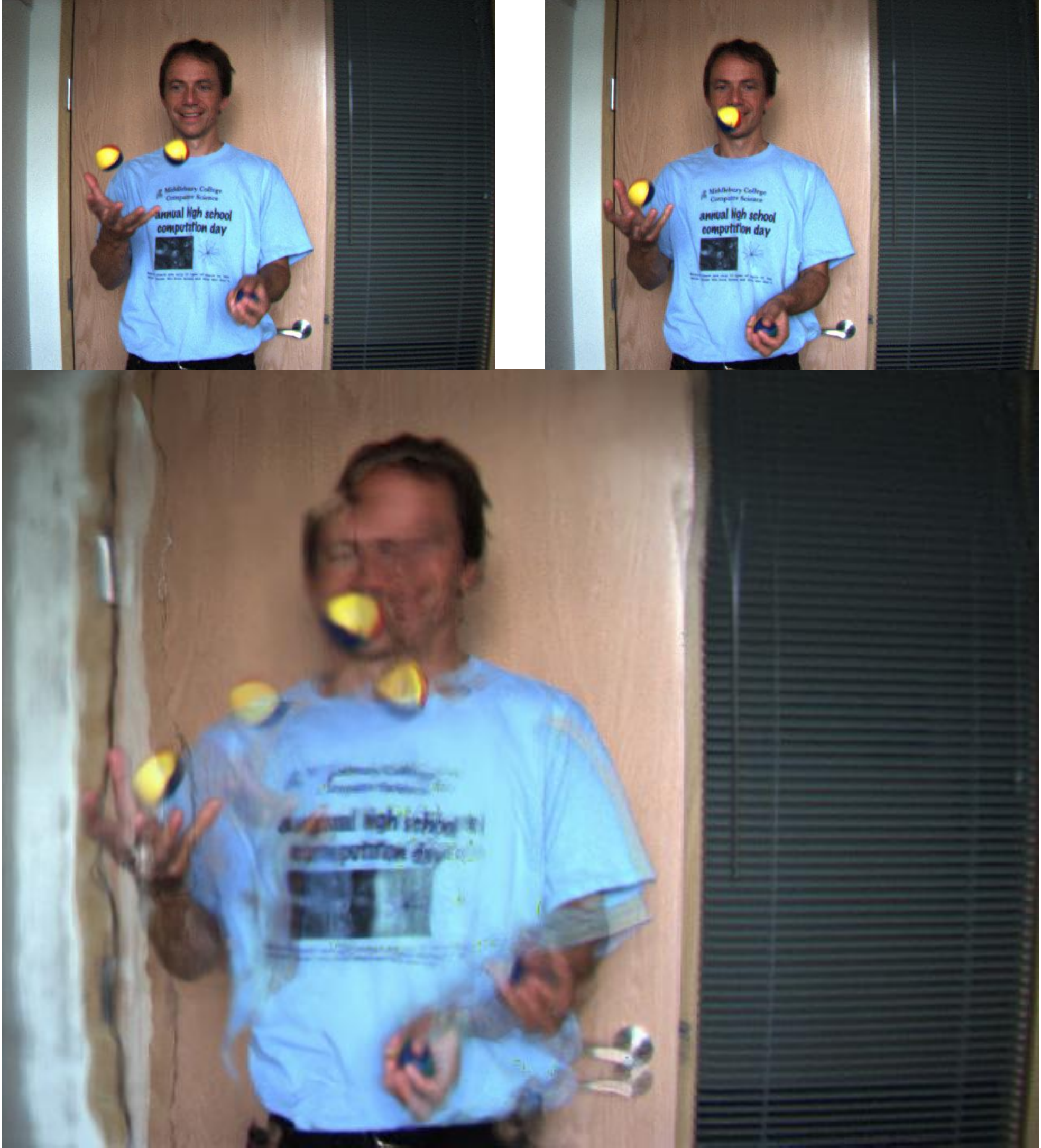}}
			\centerline{$\alpha=0.04$}
		\end{minipage}
  		\begin{minipage}[b]{0.19\linewidth}
			\centering
			\centerline{\includegraphics[width=1.0\linewidth]{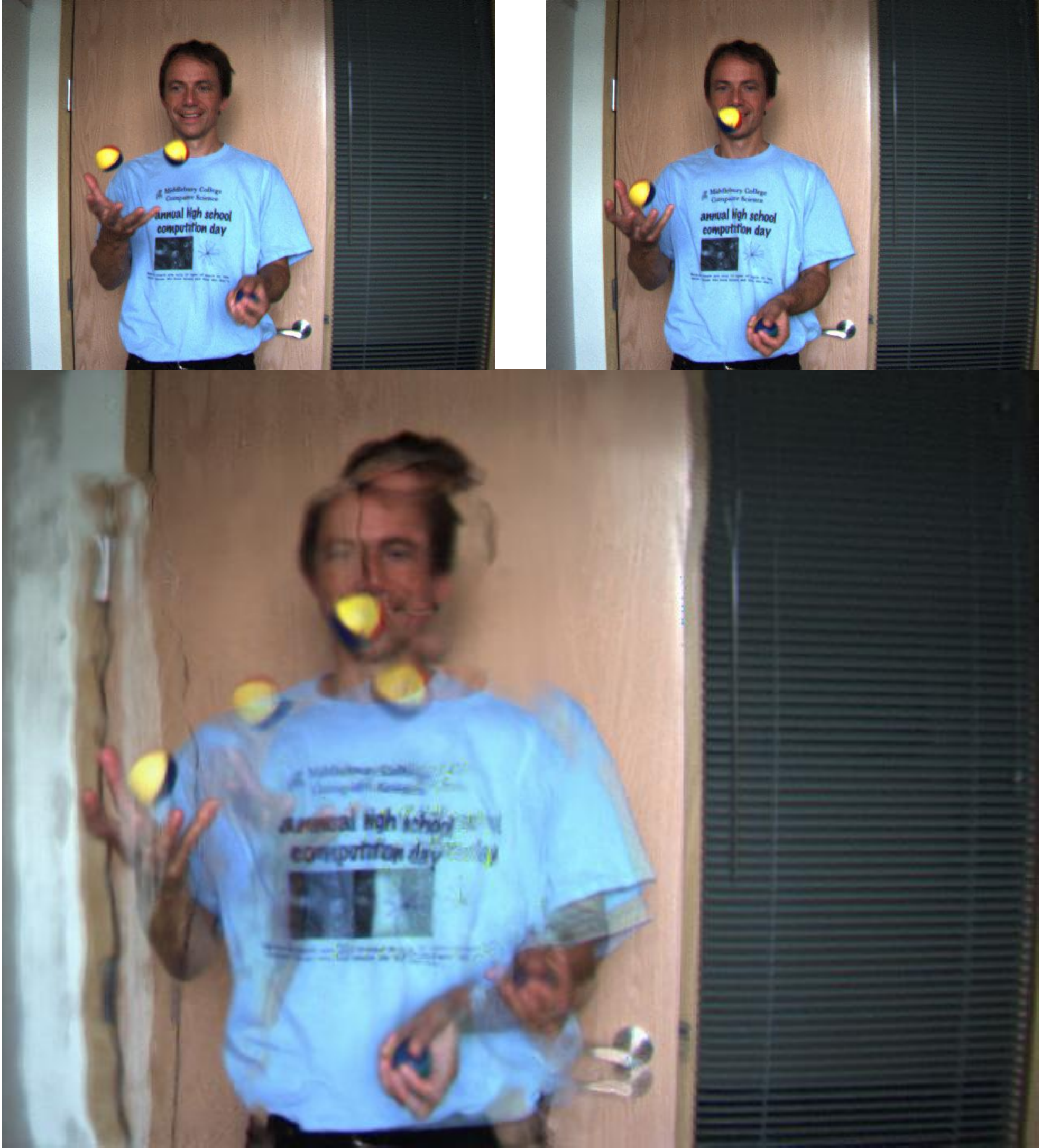}}
			\centerline{$\alpha=0.08$}
		\end{minipage}
	\end{center}
	\caption{Visual comparison of intermediate frames for the IAA attacked inputs for $\alpha \in \{0.01, 0.02, 0.04, 0.08\}$ on Middlebury dataset. The bottom is the output (intermediate frame), and the top two frames are the input frame pair in each case.}
	\label{fig:visual_comparison_alpha}
\end{figure*}

However, even if the attack achieves great performance on VIF models, the attack cost appears to be unacceptable in application scenarios due to number of frames in the videos. Thus, we proposed a novel attack for VIF models to improve attack efficiency. We find that the difference between consecutive frames in the video are very small and the corresponding adversarial perturbations are similar as well. Figure \ref{fig:visual_motivation} shows the visualization of consecutive frames in the video and the perturbations generated by the aforementioned attack. Motivated by that, instead of initializing the adversarial perturbations for each intermediate frame as zero, we propose to initialize the perturbations as the ones for the previous frames while reduce the iteration of attack. Our experiments show that our improved attack method can greatly reduce the time to generate adversarial samples under the premise of achieving the same attack performance. Besides, we use targeted attack to pursue more destructive visual attack performance and extend the attack from VIF models to higer level tasks such as video recognition models. 

Our main contributions are summarized as follows:
\begin{itemize}
    \item We apply adversarial attack to VIF models. Then we investigate a comprehensive evaluation of the adversarial robustness of the VIF models. We adopt various advanced video interpolation models based on deep learning, such as QVI\cite{QVI}, CAIN\cite{CAIN}, AdaCoF\cite{AdaCoF}, to evaluate the vulnerability of video interpolation models against adversarial attacks.
    \item We propose Inter-frame Accelerate Attack (IAA) for VIF models to improve the attack efficiency. The experiments show that our proposed attack  can accelerate the attack to generate adversarial samples while achieving comparable attack performance at the same time.
    \item We further explore target attack for VIF models and extend our method to higher level vision scenario like video recognition models. The extended experiments shows the effectiveness of our proposed method.
\end{itemize}

\section{Related Work}
\subsection{Adversarial Attack}
Recent studies have researched how to generate adversarial examples for multiple tasks such as image classification and super-resolution. Szegedy et al. \cite{Szegedy} first propose an optimization-based attack algorithm which shows that adversarial examples generating by adding a small amount of perturbation to the original images can fool CNNs successfully. Goodfellow et al. \cite{Goodfellow} show the fast sign gradient sign method (FGSM) which shows a great performance by using the sign of gradients of the model. Kurakin et al. \cite{kurakin} further develop an iterative vserion of FGSM called I-FGSM which shows a higher performance than FGSM. Madry et al. \cite{Madry} investigate a gradient-based method named projected gradient descent (PGD). Moosavi-Dezfooli et al. \cite{moosavil} study the universality of adversarial examples and propose the universal attack based on images. Liu et al. \cite{LiuAttack} show the transferability of adversarial images and developed an ensemble-based algorithm. While most works are for higer level vision task, \cite{choi2019evaluating} evaluates the adversarial robustness of the super-resolution model which is a lower level vision task. \\
Besides, when all these studies above are based on images, Wei et al. \cite{Wei} extend adversarial attack to videos by using the temporal propagation of perturbations. Chen et al.\cite{chen2021appending} propose appending adversarial frames for video recognition task. And Chen and Wei et al.\cite{chen2022attacking} suggest a bullet-screen comments adversarial frame for specific videos. In this paper, we propose a different attack method on video interpolation which is a lower level vision task compared to other video recognition, to accelerate the generation process of adversarial examples.

\subsection{Video Frame Interpolation}
Video Frame interpolation has been widely studied for a long time. Long et al. \cite{Long} propose a CNN to synthesize the intermediate frame directly. And Meyer et al. \cite{Meyer} show a phase-based video interpolation approach to combine all of the phase information. Liu et al. \cite{Liu} develop a 3D optical flow across space and time to generate the intermediate frame. Other than rely on optical flow, Niklaus et al. \cite{Niklaus01,Niklaus02} study the kernel-based methods to synthesize pixels for intermediate frames from a large neighborhood. Then Bao et al. \cite{DAIN} further combine the flow-based and kernel-based approaches to achieve a more considerable performance. Xu et al. \cite{QVI} take acceleration information into consideration so that the network can perform better in large-motion condition. To handle complex motion in videos, Lee et al. \cite{AdaCoF} propose adaptive collaboration of flows. Besides, Choi et al. \cite{CAIN} introduce channel attention to video interpolation task which performs very well. Ding et al. \cite{CDFI} investigate a compression-driven design for video interpolation networks and implement it based on \cite{AdaCoF}. Other than focus on the size of network, Kalluri et al. \cite{FLAVR} propose 3D space-time convolutions to enable end-to-end learning and inference which largely improved the efficiency of video interpolation. However, the adversarial robustness of these VIF models is not investigated yet.

\section{Methodology}
In this section, we introduce the adversarial attack (Basic Attack) on VIF models and our Inter-frame Accelerate Attack (IAA) method for generating adversarial examples on VIF models. Our proposed method is based on the basic attack and it can improve attack efficiency evidently.

\subsection{Basic Attack}
In order to make video frame interpolation models fail to generate high quality frames, we develop an attack algorithm based on Projected Gradeient Descent (PGD)\cite{Madry} method, which is one of the most effective adversarial attacks for image tasks. For video frame interpolation model, we customize PGD attack by generating adversarial perturbation pairs for the previous and next one frame of the intermediate frame. 

Let $I_i$ denote the $i$-th frame pair of the video and the corresponding attacked frame pair $\hat{I_i}$, and each frame pair contains two frames which are used to synthesis the intermediate frame. From these frames and video frame interpolation models $f(\cdot)$, we obtain intermediate frames $f(I_i)$ and $f(\hat{I_i})$. Then our goal is to maximize the loss between ground-truth $I_{gt}$ and attacked frames. We can describe the problem as the following function:
    \begin{equation}
        L(I_{gt},\hat{I_i})=\Vert f(\hat{I_i})-I_{gt}\Vert_2.
    \end{equation}
Then we apply the PGD algorithm to generate $\hat{I_i}$, which maximizes $L(I_{gt},\hat{I_i})$ with the $l_\infty$-norm constraint. In the process, we update the perturbation iteratively added into original images which is denoted as $\widetilde{P}_n$ as the following function:
    \begin{equation}
        \label{eq2}
        \widetilde{P}_{n+1}=eps*sgn(\nabla L(I_{gt},\hat{I}_n))
    \end{equation}
    where $eps$ represents the amount of perturbation generated by each iteration and $sgn(\nabla L(I_{gt},\hat{I}_n))$ calculates the gradient of $L(I_{gt},\hat{I}_n)$. And $n$ denotes the $n$-th iteration.
    \begin{equation}
        P_{n+1} =clip_{-\alpha,\alpha}(clip_{0,1}(\widetilde{P}_{n+1}+\hat{I}_n)-I_{gt}). 
    \end{equation}
    The parameter $\alpha$ limits the amount of perturbation added into original images so that we can ensure the perturbation is invisible. And $clip_{a,b}(\cdot)$ can be defined as
    \begin{equation}
        clip_{a,b}(I)=min(max(I,a),b).
    \end{equation}
    By iteratively updating $P_i$, we can obtain the final adversarial example by:
    \begin{equation}
        \hat{I_i}=I_0+P_T
    \end{equation}
where $T$ is the number of iterations.

\begin{algorithm}
    \caption{Inter-frame Accelerate Attack}
    \label{fast_video_attack}
    \begin{algorithmic}[1]
    \REQUIRE model $f(\cdot)$, input frame pairs $I$, maximum attack iteration $T$, step size $eps$, perturbation boundary $\alpha$, and frame pair number $N$
    \ENSURE adversarial example $\hat{I}$
    \STATE $P_0 \gets 0$, $\hat{I}_0 \gets I$   \\
    $\#$ attack each frame pair in the video
    \FOR{$i \gets 0$ to $N-1$} 
    {
        \IF{$i = 0$}
            \STATE $P_0 \gets 0$
        \ELSE 
            \STATE $P_i \gets P_{i-1}$
        \ENDIF
        $\#$ perform attack for $T/2$ iterations
        \FOR{$t \gets 0$ to $T/2$}{
            \STATE $L(I_{gt},\hat{I_i}) \gets \Vert f(\hat{I_i})-I_{gt}\Vert_2$
            \STATE $P_i \gets clip_{-\alpha, \alpha}(eps*sgn(\nabla L(I_{gt},\hat{I}_i)))$
            \STATE $\hat{I}_i \gets \hat{I}_i + P_i$ 
        }
        \ENDFOR
    }
    \ENDFOR
    \end{algorithmic}
\end{algorithm}

\subsection{Inter-frame Accelerate Attack}

\textbf{Non-targeted Attack.} Although the basic attack (BA) method can generate imperceptible adversarial perturbations for video interpolation task, it costs too much time and computation resources to generate adversarial examples for each frame in the video. And unlike video recognition task which only need to mislead the model to output negative labels, to attack VIF models, we have to destroy every frame in the video which disables other video attacks. We propose a new method named Inter-frame Accelerate Attack (IAA) to accelerate the attack process. The intuition here is to make full use of the similarity between consecutive frames in the video. Due to the temporal continuity of adjacent frames in the video and the difference between them is often very small, it is possible for us to generate $P_i$ based on $P_{i-1}$. Algorithm \ref{fast_video_attack} shows the process of our proposed IAA method.

Let $P_i,i \in \{0,1,\dots,N-1\}$ denote the perturbation added to $i$-th input frame pair. In basic attack method, we set up the initial value for $P_i$ before the first iteration as zero.  As can be seen in Algorithm \ref{fast_video_attack}, in IAA, we set the initial value as
    \begin{equation}
        \begin{cases}
        P_i=0,&i=0 \\ P_i=P_{i-1},&i\neq0
        \end{cases}
    \end{equation}
By inheriting the perturbation information from the previous frame pair, it will be possible for us to reduce the amount of iteration while reaching the same attack performance. In our experiment, we halve the number of the attack iterations for input frames other than the first frame. By this way, we can save almost 50\% attack time in the video frame interpolation task theoretically.

\textbf{Targeted Attack.} Although the basic attack and Inter-frame Accelerate Attack for VIF models can degrade the quality of generated intermediate frames, the deterioration is measured by PSNR and SSIM. We want to further explore the attack method to degrade the visual quality of intermediate frames. Targeted attack in image classification task aims to mislead the classifier to specific labels. And in super-resolution task, the targeted attack is to make the model generate images that are more similar to the target than the original ground-truth \cite{choi2019evaluating}. Here, we apply targeted attack to VIF models so that the generated intermediate frame can be more similar to the target. To make that, we simply modify the Eq. \ref{eq2} as 
    \begin{equation}
        \widetilde{P}_{n+1}=-eps*sgn(\nabla L(I_{t},\hat{I}_n))
    \end{equation}
where $I_{t}$ is the target image.

\subsection{Attack Transferability to Video Recognition Models}
Video frame interpolation task is a lower level vision task compared to video recognition task. We further extend our attack method to higher level vision tasks such as video recognition task. For this, we divide the frames in a single video to multiple frame groups $G$, where the number of frames in $G$ is more than the minimum required input frames for the video recognition model. Our goal is to fool the model to misclassify the video by generating adversarial perturbations based on gradients. For each group, we obtain the perturbations by:
    \begin{equation}
        {P}_{n+1}= {P}_{n} + eps*sgn(\nabla L(1_{G}, f(\hat{G}))
    \end{equation}
where $1_{G}$ is the ground-truth label of the video, $n$ denotes the $n$-th attack iteration and perturbation $P$ is limited within [$-\alpha, +\alpha$], the same as Algorithm \ref{fast_video_attack}. Instead of initializing the perturbations as zeros, we use the perturbation values of the previous group which is from the same video as the initialization value.
    \begin{equation}
        {P}_0^i= P_T^{i-1}
    \end{equation}
where $i$ denotes the $i$-th group frames in the video and $T$ denotes the maximum iteration of $(i-1)$-th frame group.

 \begin{figure*}[t]
	\begin{center}
		\centering
		\begin{minipage}[b]{0.24\linewidth}
			\centering
			\centerline{\includegraphics[width=1.0\linewidth]{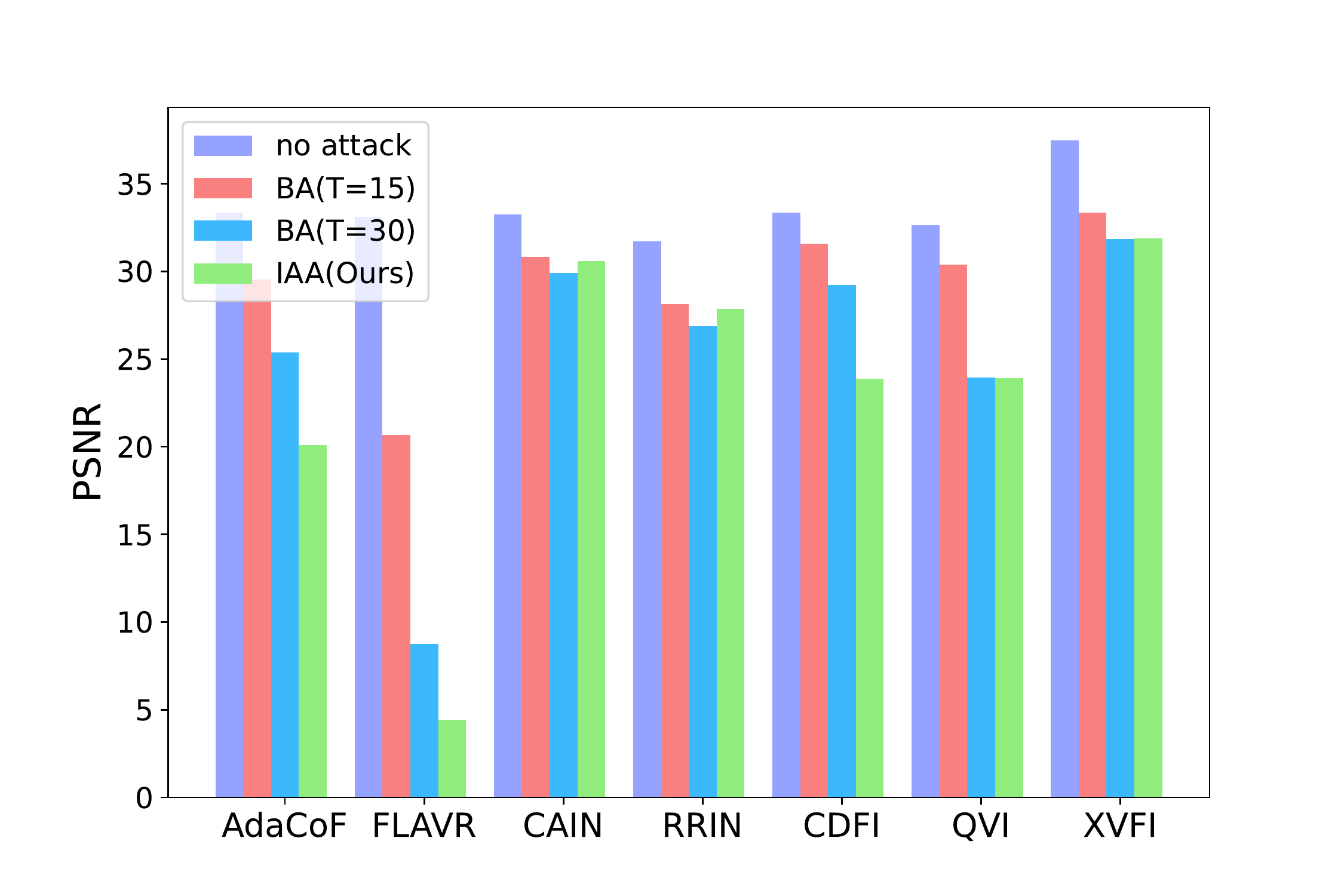}}
			\centerline{(a) UCF-101, $\alpha=0.01$}
		\end{minipage}
		\begin{minipage}[b]{0.24\linewidth}
			\centering
			\centerline{\includegraphics[width=1.0\linewidth]{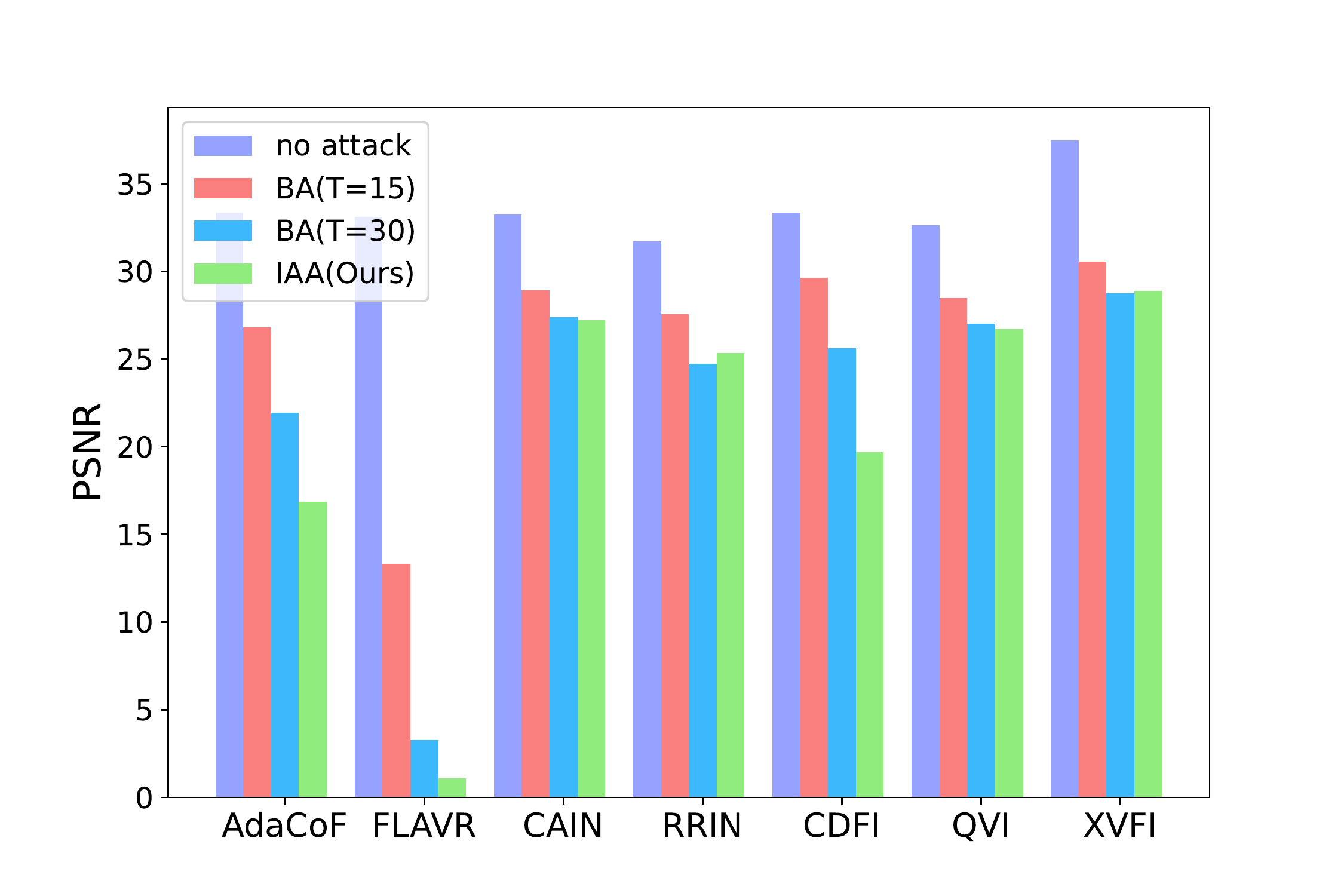}}
			\centerline{(b) UCF-101, $\alpha=0.02$}
		\end{minipage}
		\begin{minipage}[b]{0.24\linewidth}
			\centering
			\centerline{\includegraphics[width=1.0\linewidth]{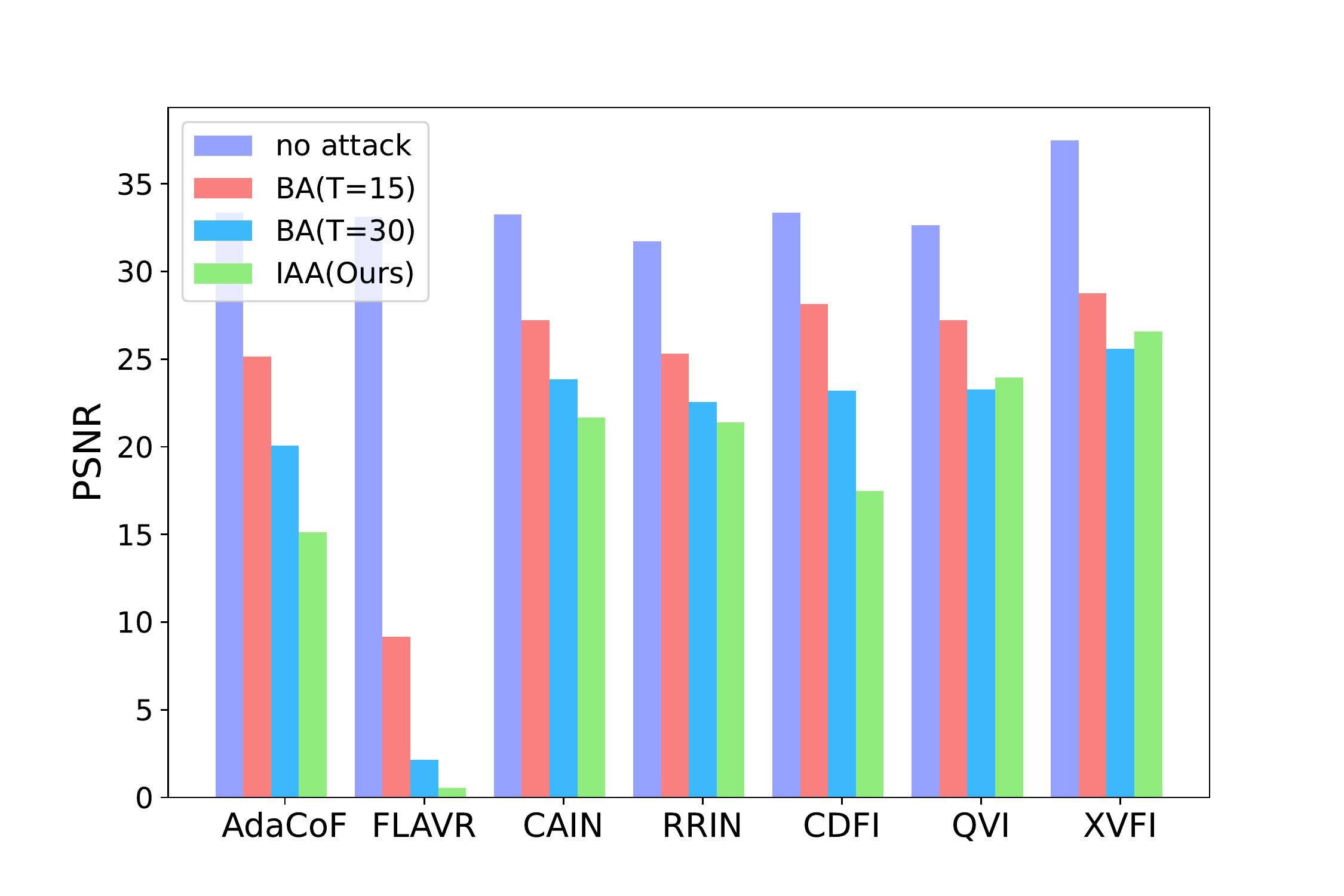}}
			\centerline{(c) UCF-101, $\alpha=0.04$}
		\end{minipage}
		\begin{minipage}[b]{0.24\linewidth}
			\centering
			\centerline{\includegraphics[width=1.0\linewidth]{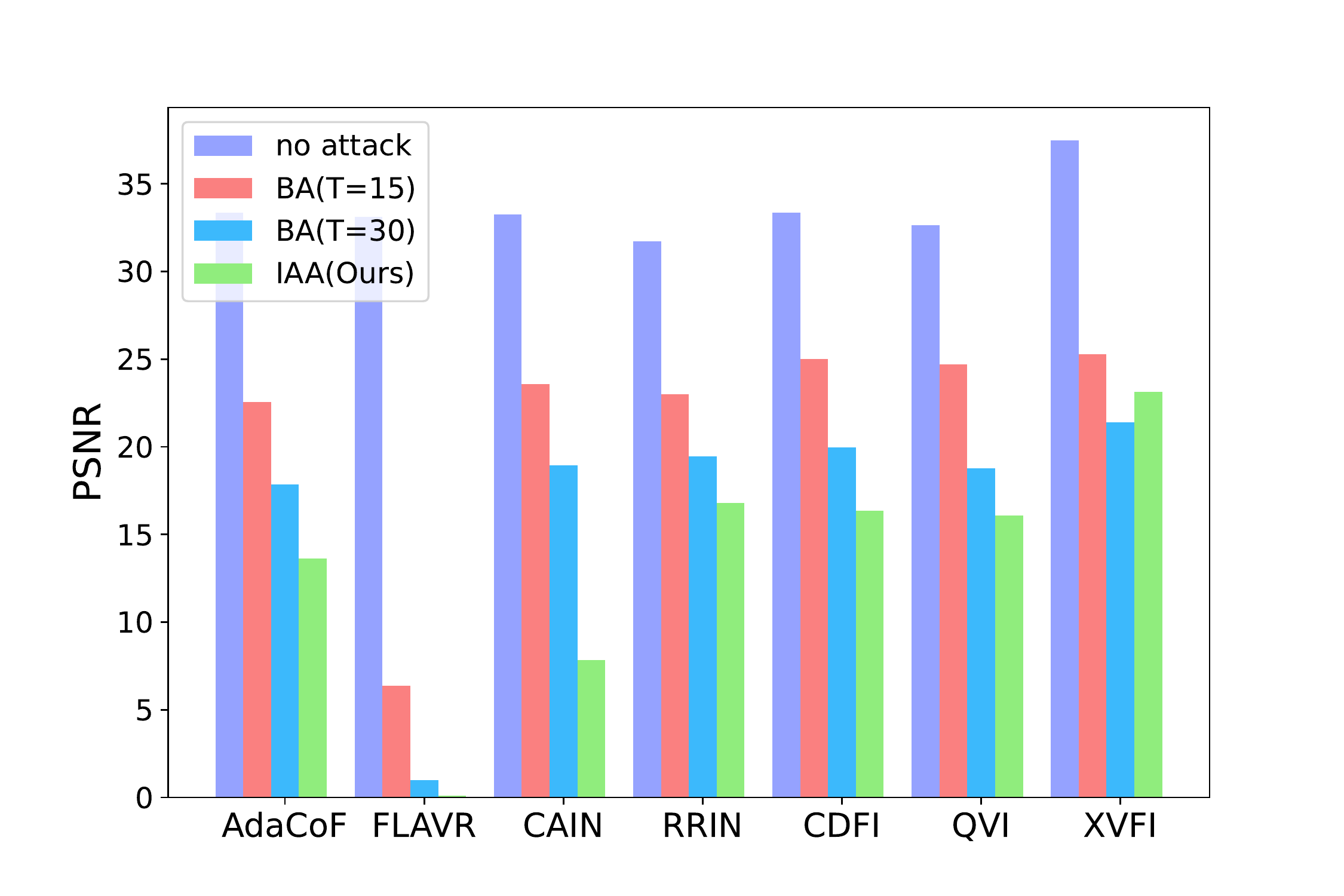}}
			\centerline{(d) UCF-101, $\alpha=0.08$}
		\end{minipage}
  		\begin{minipage}[b]{0.24\linewidth}
			\centering
			\centerline{\includegraphics[width=1.0\linewidth]{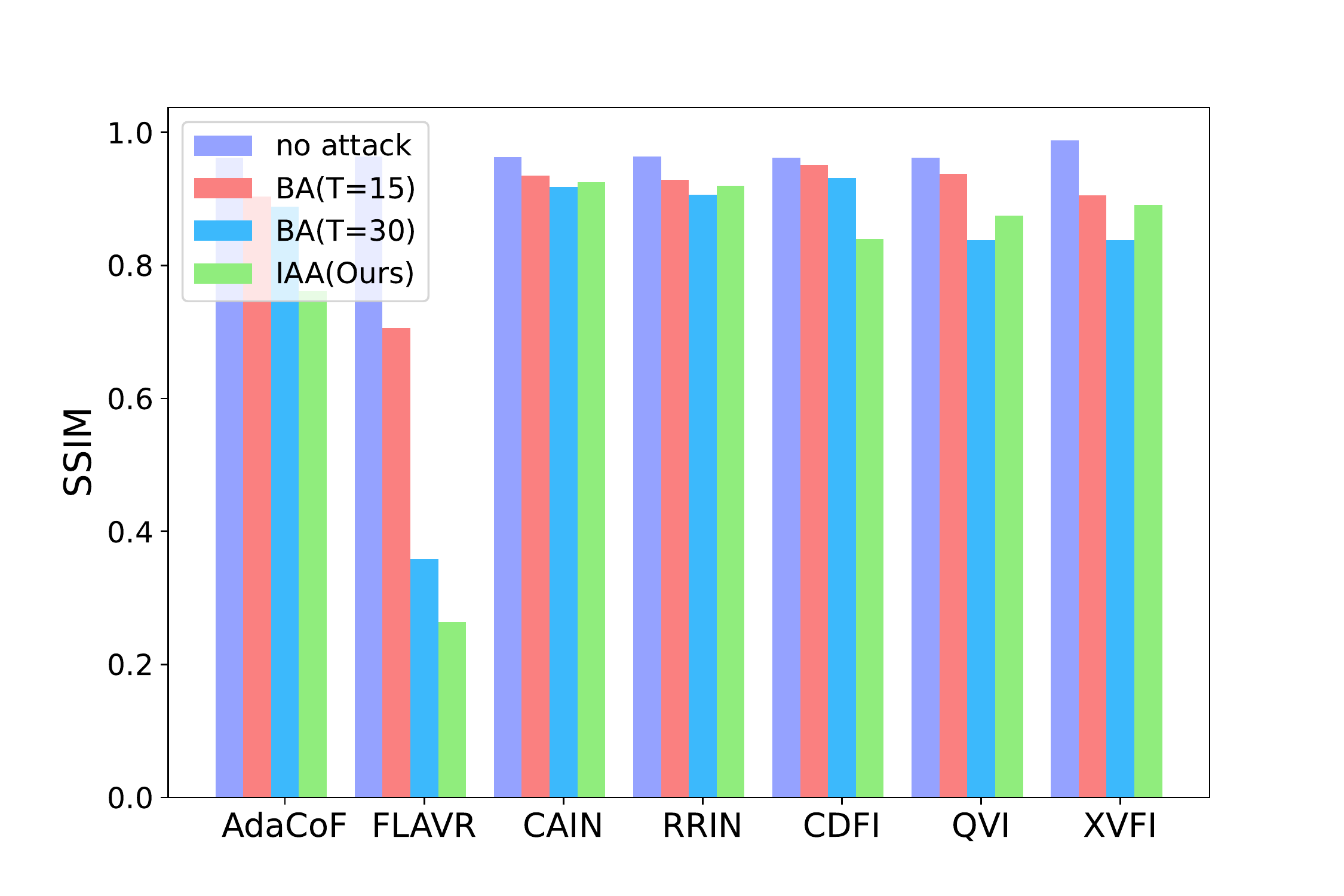}}
			\centerline{(e) UCF-101, $\alpha=0.01$}
		\end{minipage}
		\begin{minipage}[b]{0.24\linewidth}
			\centering
			\centerline{\includegraphics[width=1.0\linewidth]{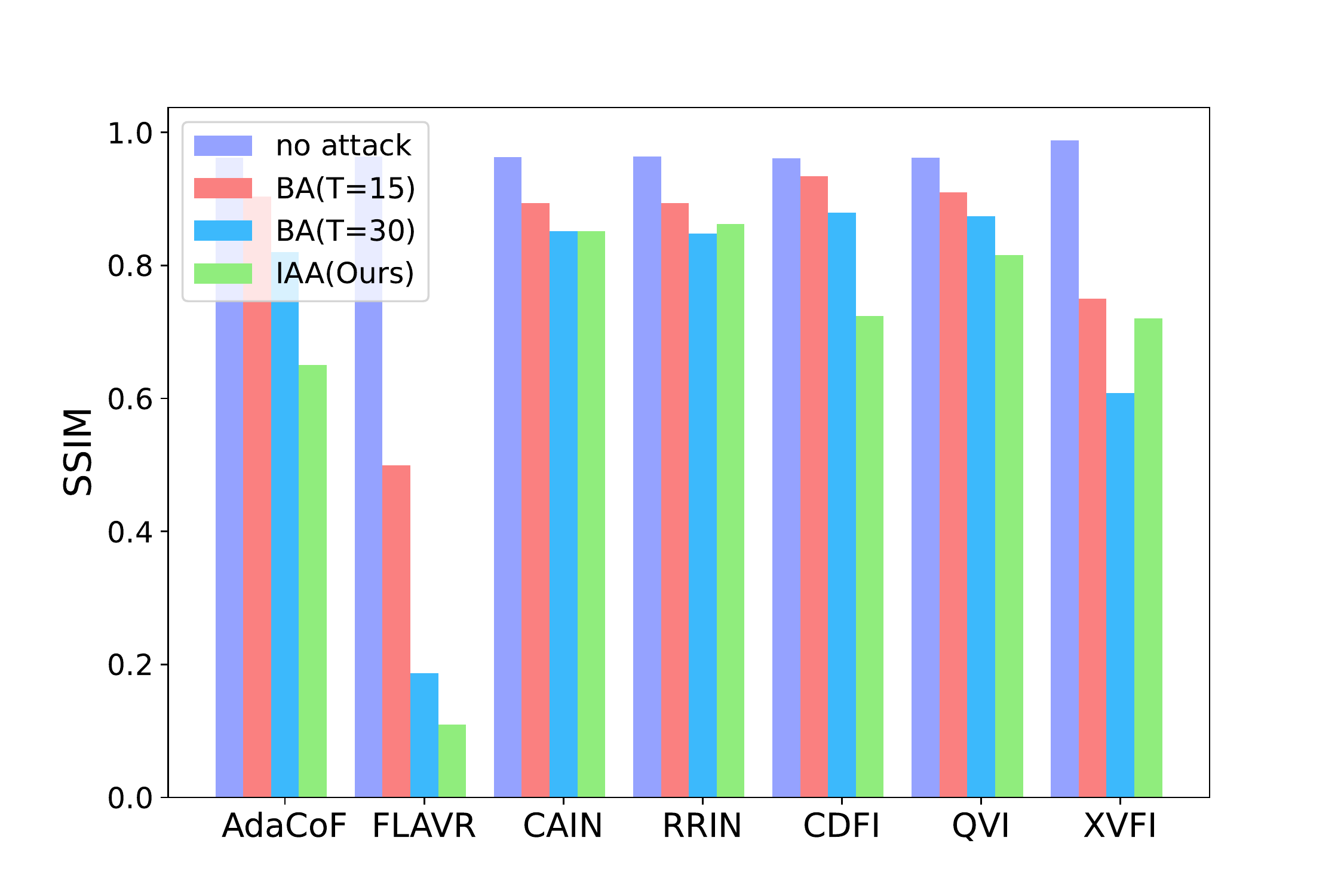}}
			\centerline{(f) UCF-101, $\alpha=0.02$}
		\end{minipage}
		\begin{minipage}[b]{0.24\linewidth}
			\centering
			\centerline{\includegraphics[width=1.0\linewidth]{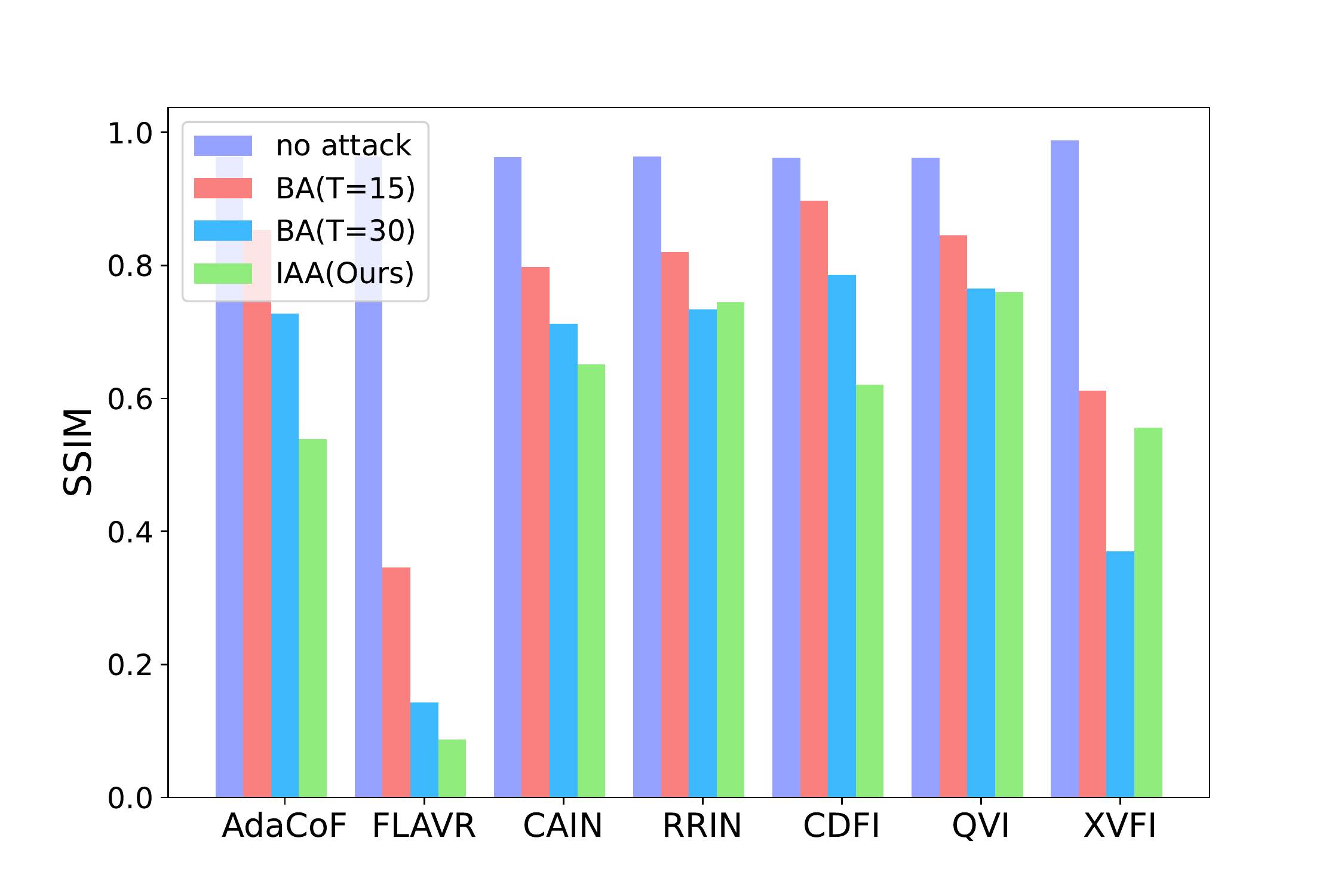}}
			\centerline{(g) UCF-101, $\alpha=0.04$}
		\end{minipage}
		\begin{minipage}[b]{0.24\linewidth}
			\centering
			\centerline{\includegraphics[width=1.0\linewidth]{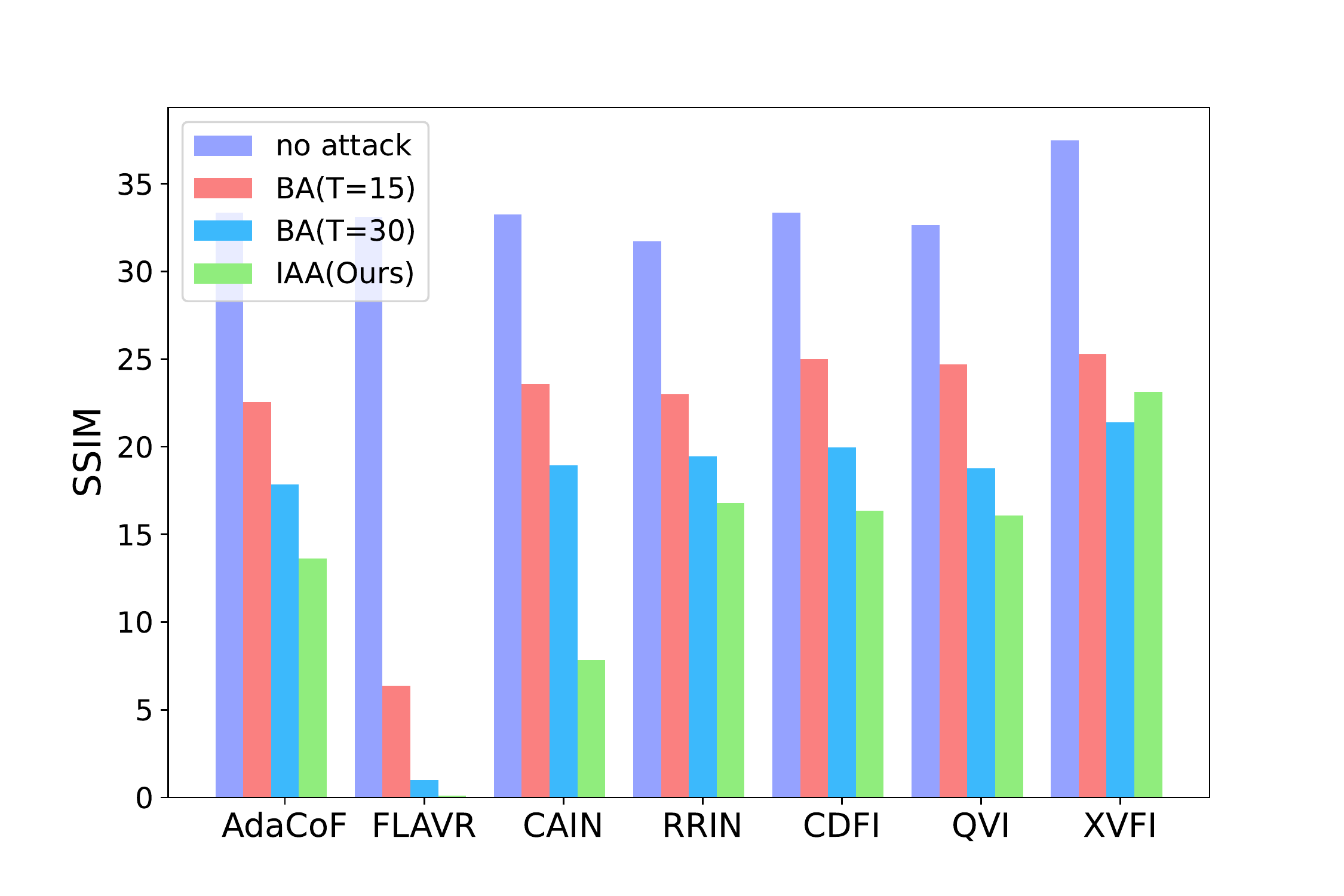}}
			\centerline{(h) UCF-101, $\alpha=0.08$}
		\end{minipage}
	\end{center}
	\caption{Comparison of the PSNR and SSIM values of generated intermediate frames of different $\alpha$ on UCF-101 dataset.}
	\label{fig:psnr_and_ssim_comparison}
\end{figure*}

\begin{figure*}[t]
	\begin{center}
		\centering
		\begin{minipage}[b]{0.24\linewidth}
			\centering
			\centerline{\includegraphics[width=1.0\linewidth]{figures/middlebury_groundtruth_5.pdf}}
			\centerline{(a) Groundtruth}
		\end{minipage}
		\begin{minipage}[b]{0.24\linewidth}
			\centering
			\centerline{\includegraphics[width=1.0\linewidth]{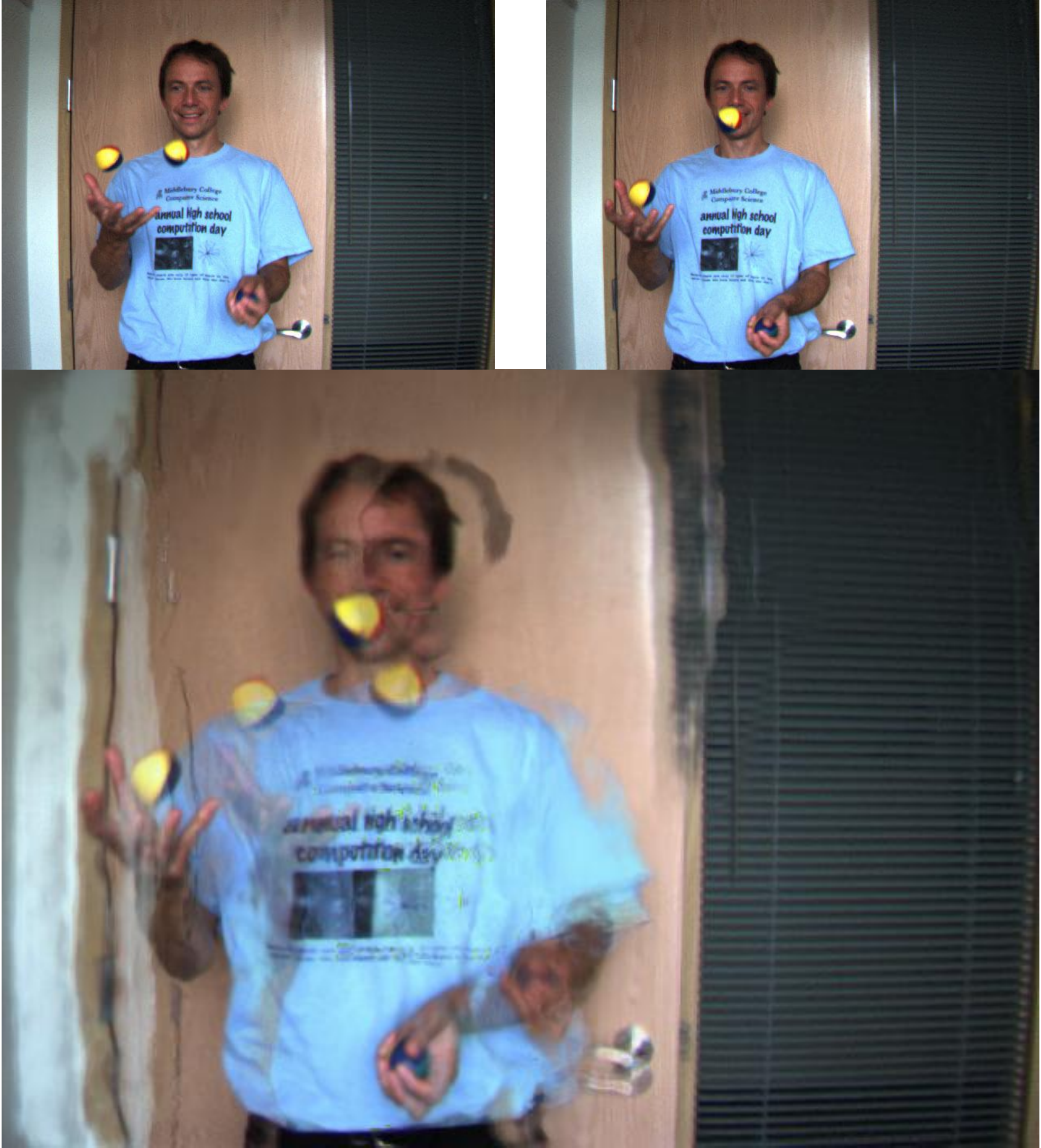}}
			\centerline{(b) AdaCoF}
		\end{minipage}
		\begin{minipage}[b]{0.24\linewidth}
			\centering
			\centerline{\includegraphics[width=1.0\linewidth]{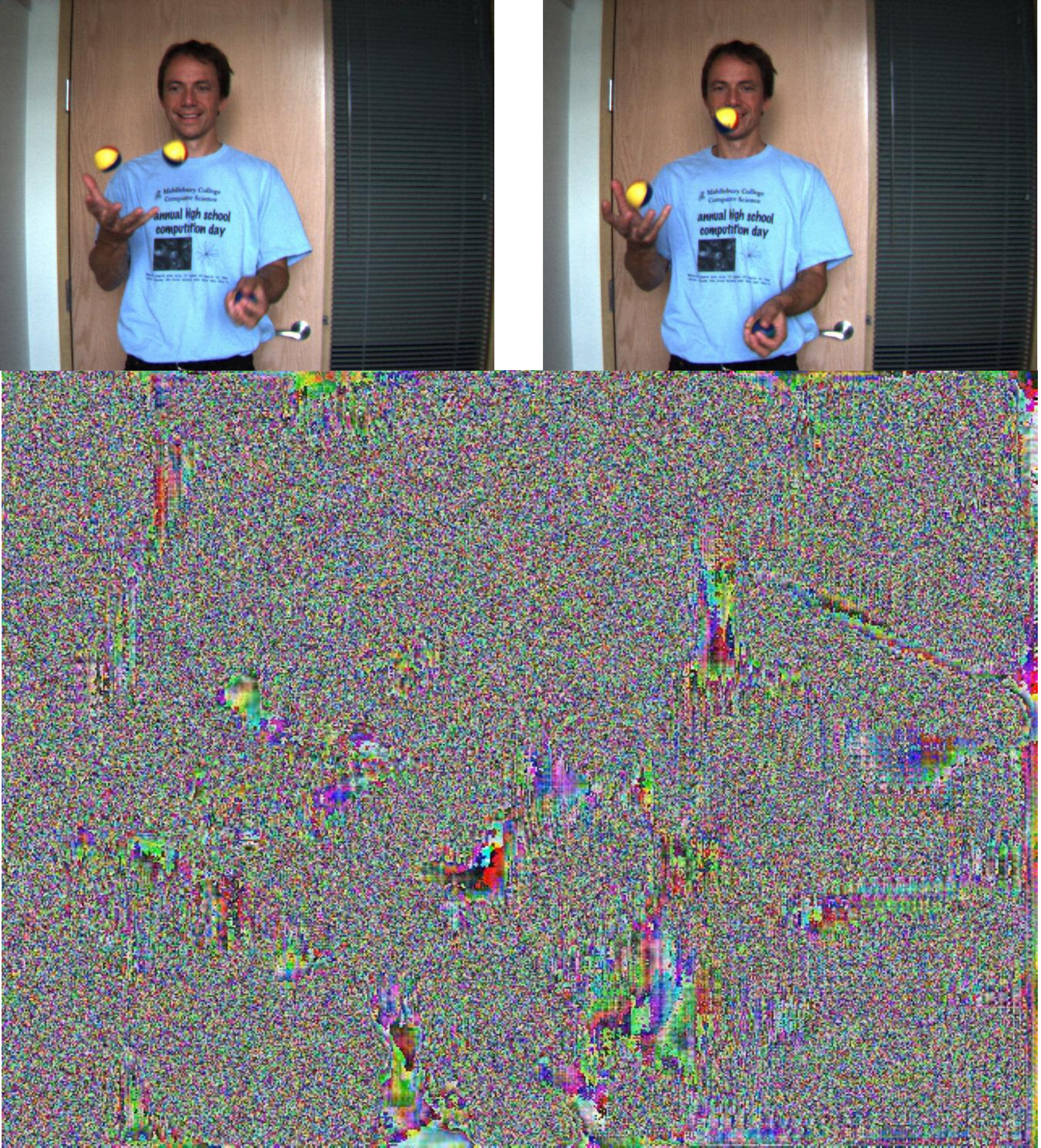}}
			\centerline{(c) FLAVR}
		\end{minipage}
		\begin{minipage}[b]{0.24\linewidth}
			\centering
			\centerline{\includegraphics[width=1.0\linewidth]{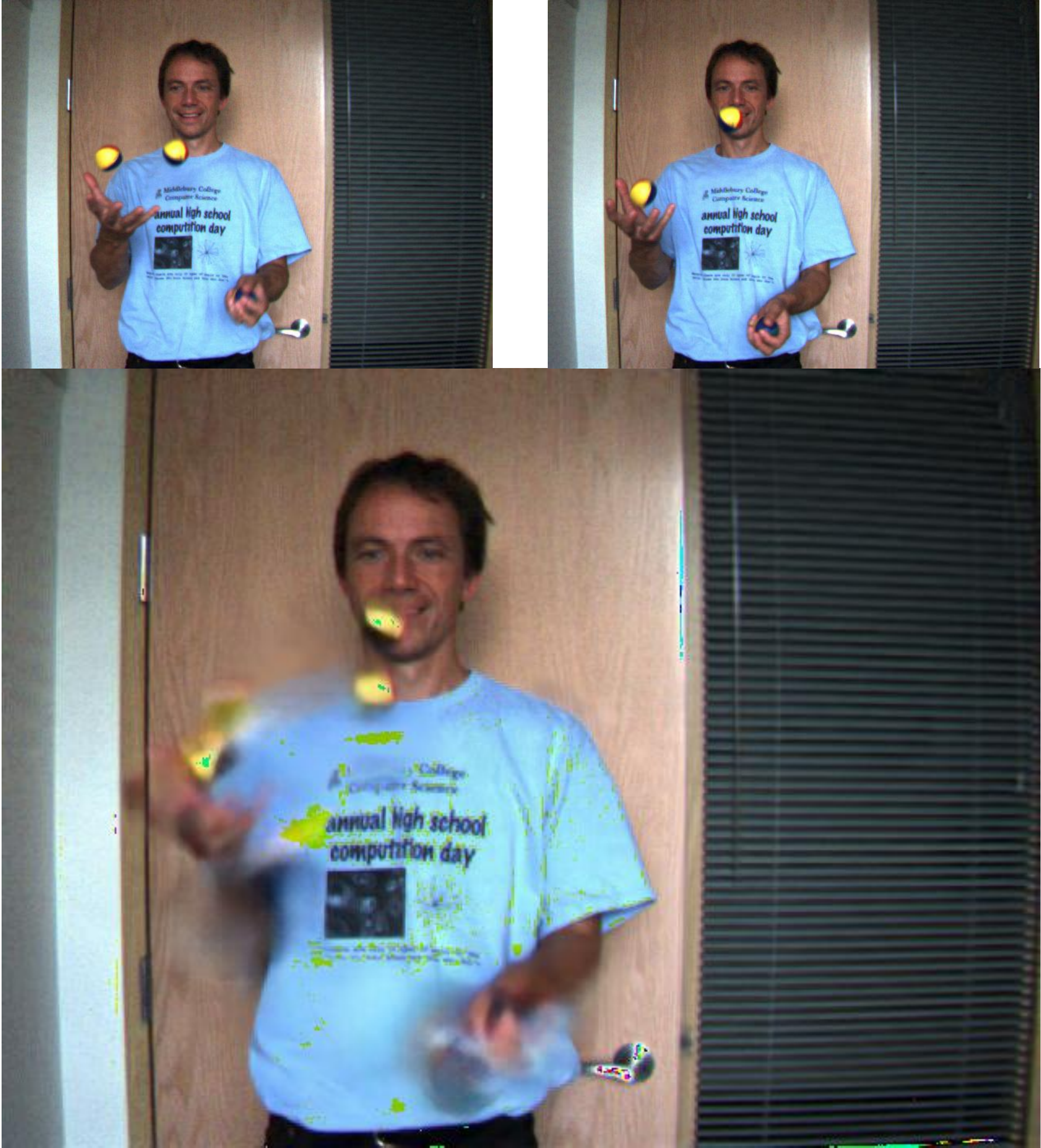}}
			\centerline{(d) CAIN}
		\end{minipage}
  		\begin{minipage}[b]{0.24\linewidth}
			\centering
			\centerline{\includegraphics[width=1.0\linewidth]{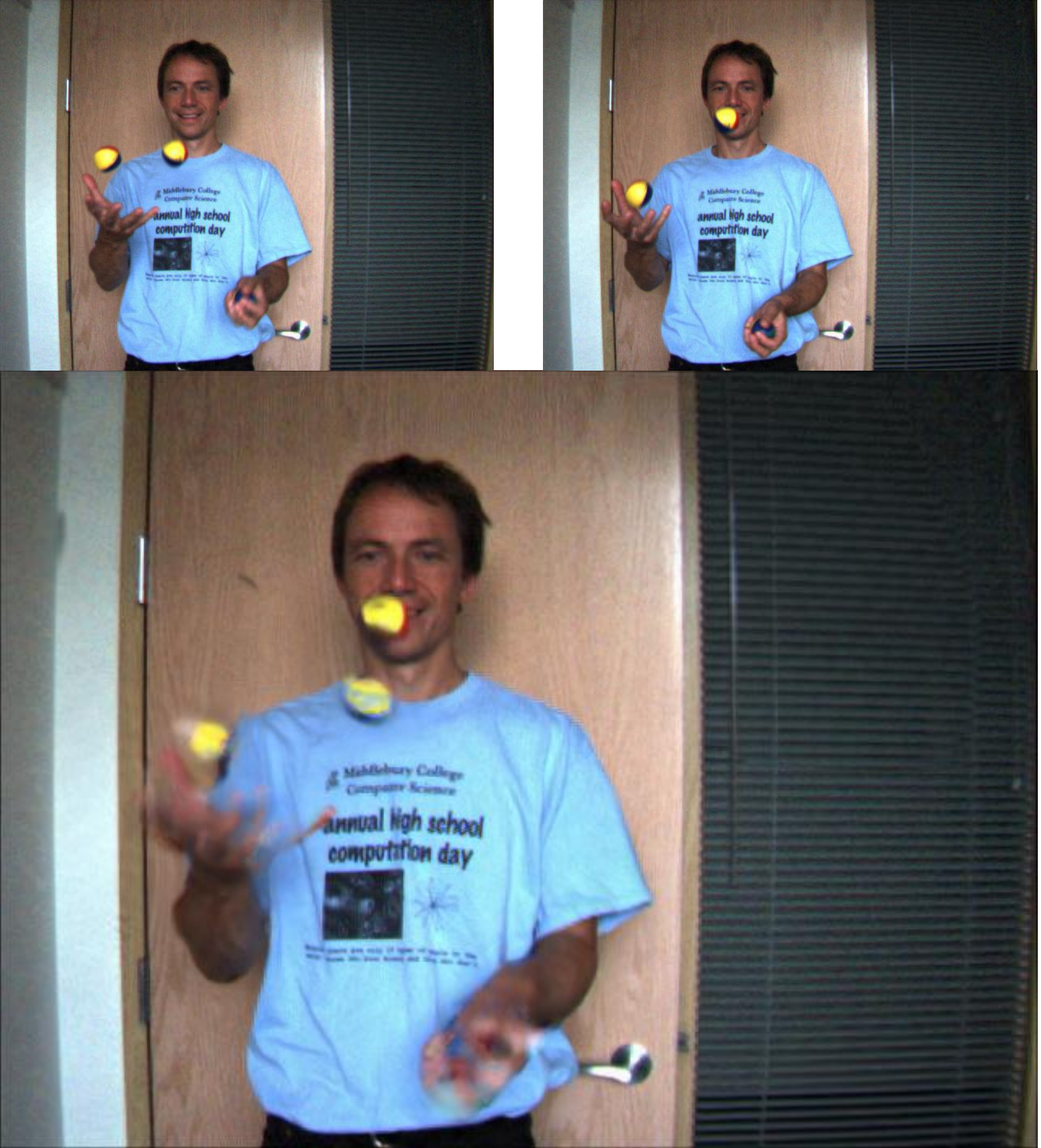}}
			\centerline{(e) RRIN}
		\end{minipage}
		\begin{minipage}[b]{0.24\linewidth}
			\centering
			\centerline{\includegraphics[width=1.0\linewidth]{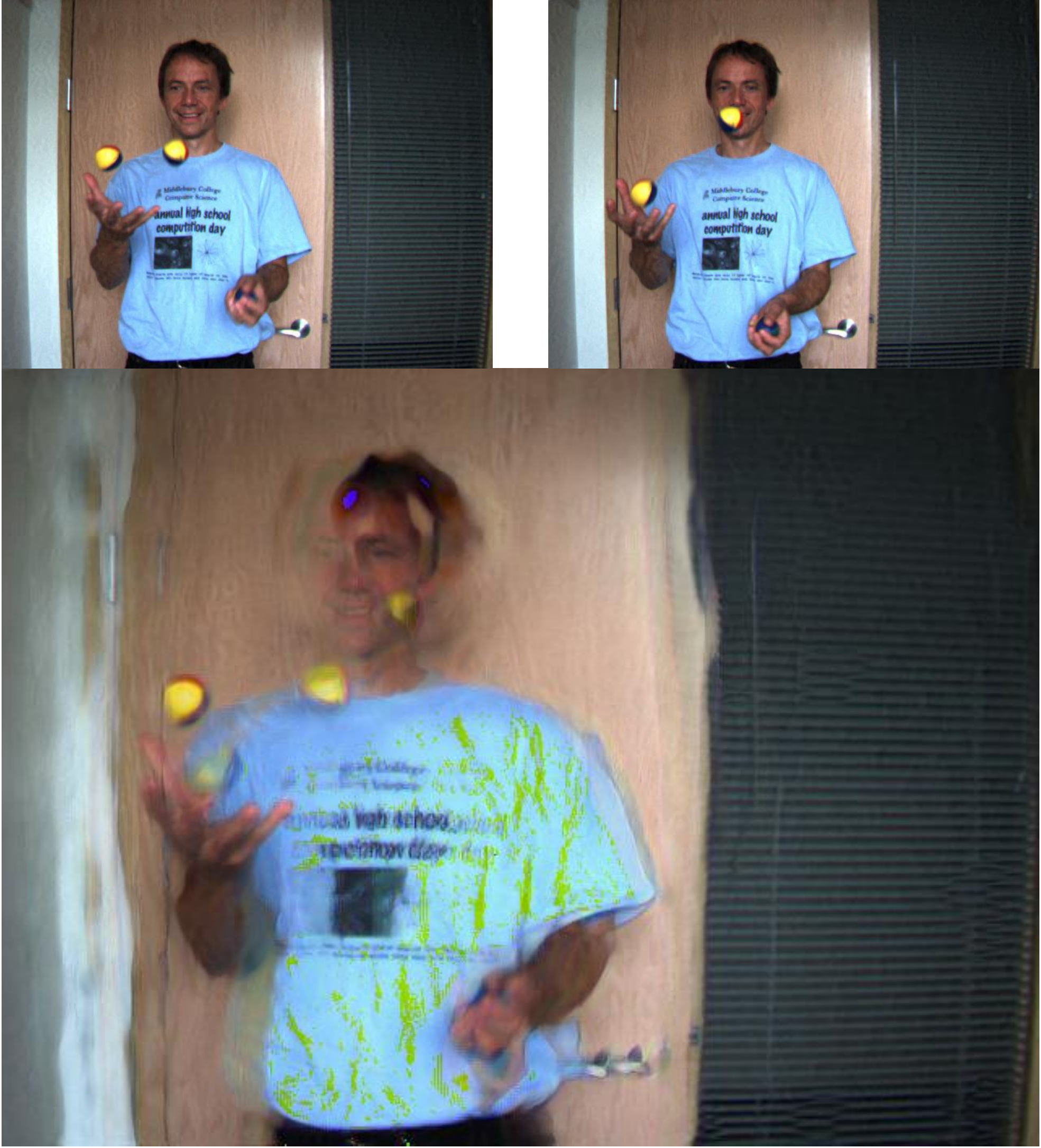}}
			\centerline{(f) CDFI}
		\end{minipage}
		\begin{minipage}[b]{0.24\linewidth}
			\centering
			\centerline{\includegraphics[width=1.0\linewidth]{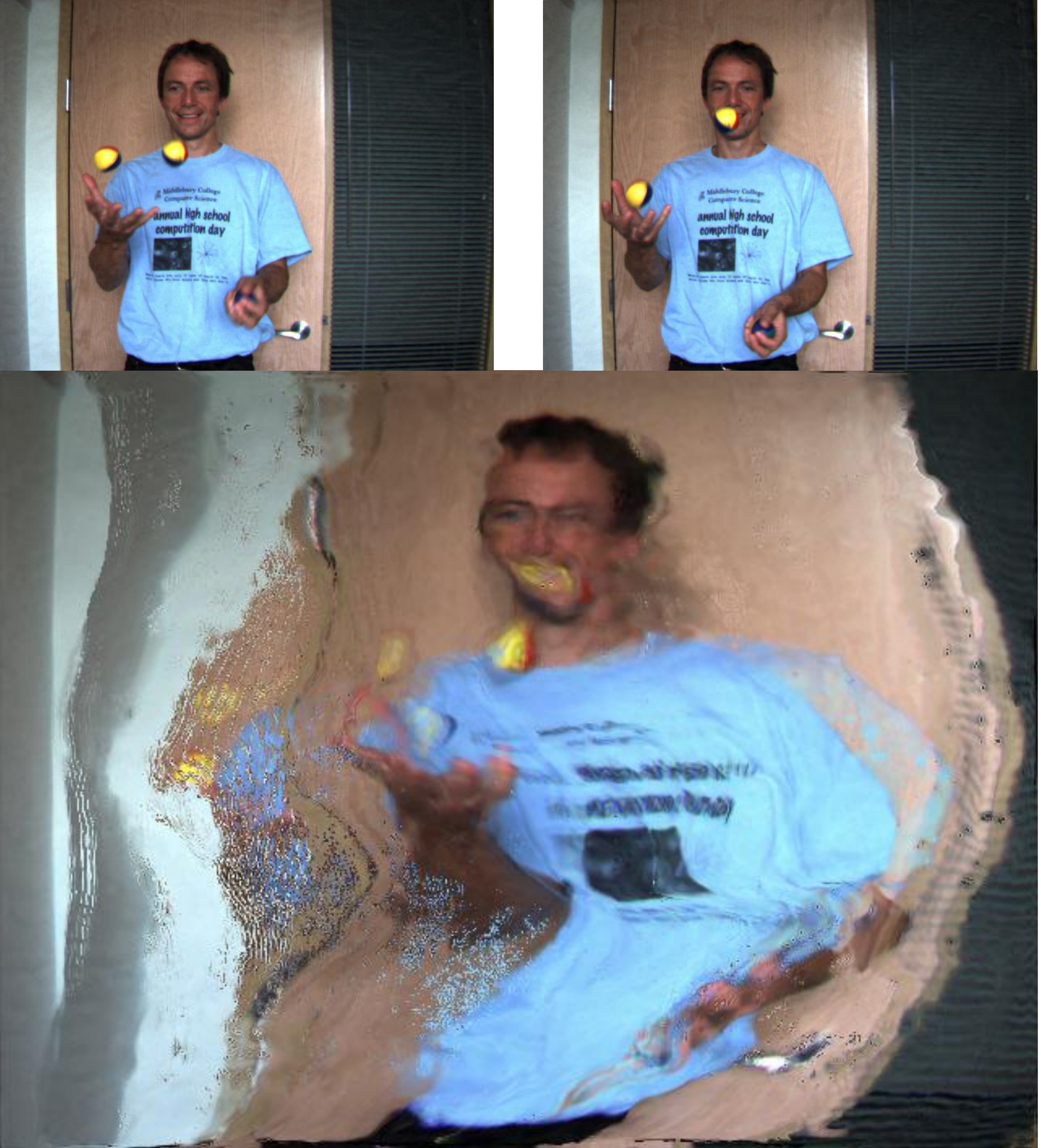}}
			\centerline{(g) QVI}
		\end{minipage}
		\begin{minipage}[b]{0.24\linewidth}
			\centering
			\centerline{\includegraphics[width=1.0\linewidth]{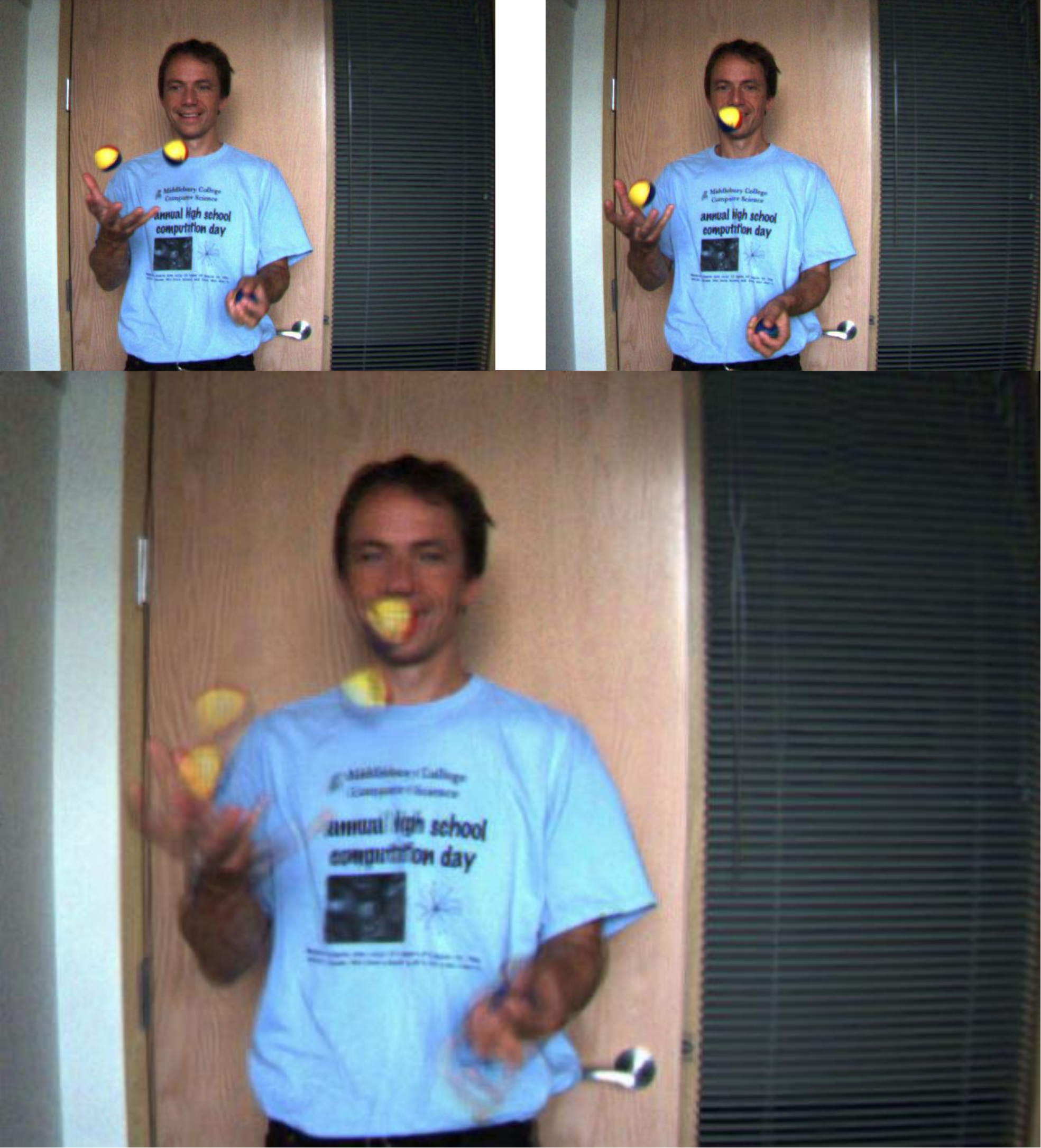}}
			\centerline{(h) XVFI}
		\end{minipage}
	\end{center}
	\caption{Visual comparison of the IAA attacked intermediate frames of $\alpha=0.02$ on Middlebury dataset. The bottom is the output (intermediate frame), and the top two frames are the input frame pair in each case.}
	\label{fig:visual_all_models}
\end{figure*}

\section{Experiments}

\subsection{Experiment Settings}
\textbf{Datasets.} We use three datasets that are widely used for video frame interpolation methods: Middlebury, Vimeo90K, and UCF-101. The UCF-101 \cite{soomro2012ucf101} used in our experiments is the standard dataset collected from Youtube, which contains 13,320 videos with 101 action classes covering a broad set of activities. For ease of our experiments, we streamline it and take the first 50 frames of the first video in the 101 categories, the dataset applied in our experiments contains 101 sets of data, each set of data has 50 consecutive frames. The Vimeo90K \cite{xue2019video} is a large-scale high-quality video dataset for lower-level video processing. Our work adopts the Septuplet part of the Vimeo90K dataset, which contains 7824 videos, each of which contains 7 consecutive frames. The Middlebury \cite{baker2011database} consists of high-resolution stereo sequences with complex geometry and pixel-accurate ground-truth disparity data. We use the Middlebury-other dataset, after excluding data containing only 2 frames. The existing dataset contains 10 groups of data, each with 9 consecutive frames.

\textbf{Metrics.} We use Peak Signal-to-Noise Ratio (PSNR), Structural Similarity (SSIM) and attack time to measure the robustness of the VIF models against our adversarial attacks. For different datasets and VIF models, we respectively calculate the PSNR/SSIM values between the ground truth and the generated intermediate frame. Then, the PSNR/SSIM values between the ground truth and the video frame attacked by the attacks are calculated. Besides, we assist in understanding the performance of basic attack and IAA by recording the attack time. For different VIF models, the attack time of basic attack method is recorded for 15 and 30 iterations respectively. In our experiment settings, it can be concluded that the time of IAA method is similar to that of basic attack method for 15 iterations, but far less than that of 30 iterations, so it can be judged that our proposed IAA method has good performance.

\begin{table}[]
	\small
	\setlength{\tabcolsep}{0.5em}
	\renewcommand{\arraystretch}{1.05}
 	\caption{Properties of the video frame interpolation models.}
	\label{vfi_methods_properties}
	\begin{center}
		\begin{tabular}{l|rrr}
			\textbf{Method} & \textbf{parameters} & \textbf{layers} & \textbf{class} \\
			\hline
			QVI \cite{QVI} & 29.2M & 81 & flow-based \\
			FLAVR \cite{FLAVR} & 42.1M & 37 & kernel-based \\
			CAIN \cite{CAIN} & 42.7M & 247 & kernel-based \\
			CDFI \cite{CDFI} & 4.9M & 128 & kernel-based \\
			AdaCoF \cite{AdaCoF} & 21.8M & 59 & kernel-based \\
			RRIN \cite{RRIN} & 19.1M & 81 & flow-based \\
                XVFI \cite{XVFI} & 5.6M & 34 & flow-based \\
		\end{tabular}
	\end{center}
\end{table}

\textbf{Video Interpolation Models.} Our experiments use seven advanced deep learning-based VIF methods with various model sizes and properties, including RRIN \cite{RRIN}, FLAVR \cite{FLAVR}, QVI \cite{QVI}, CAIN \cite{CAIN}, CDFI \cite{CDFI}, AdaCoF \cite{AdaCoF} and XVFI \cite{QVI}. QVI is a flow-based model, while FLAVR, CAIN, CDFI and AdaCoF are kernel-based. And XVFI is the first method proposed for 4K videos with large motion. In the experiments, we employ the pre-trained models provided by original authors. Table \ref{vfi_methods_properties} shows their characteristics in terms of the number of model parameters, the number of convolutional layers, and class according to \cite{dong2022video}.

\textbf{Implementation details.} For all our methods, we generate two adversarial perturbations for the previous and next frame of the intermediate frame. We set the $\alpha \in \{0.01, 0.02, 0.04, 0.08\}$ and $T \in \{15, 30\}$. And we set $eps \in \{0.0005, 0.001, 0.002, 0.003\}$  for each $\alpha$ value. Our adversarial attack methods are implemented on the pytorch framework, and running on one Nvidia V100-32GB GPU.

\begin{figure*}[t]
	\begin{center}
		\centering
		\begin{minipage}[b]{0.49\linewidth}
			\centering
			\centerline{\includegraphics[width=0.8\linewidth]{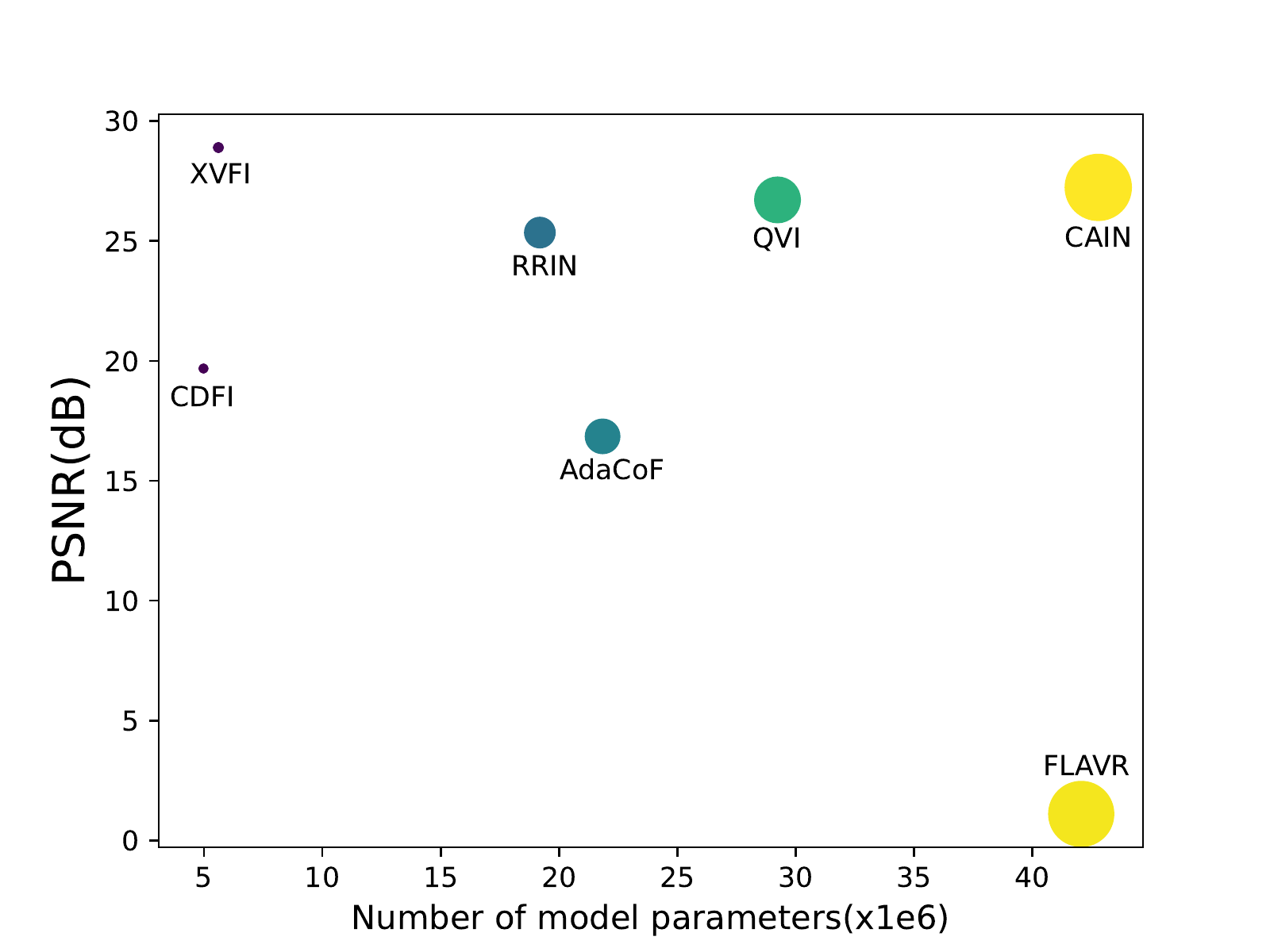}}
		\end{minipage}
		\begin{minipage}[b]{0.49\linewidth}
			\centering
			\centerline{\includegraphics[width=0.8\linewidth]{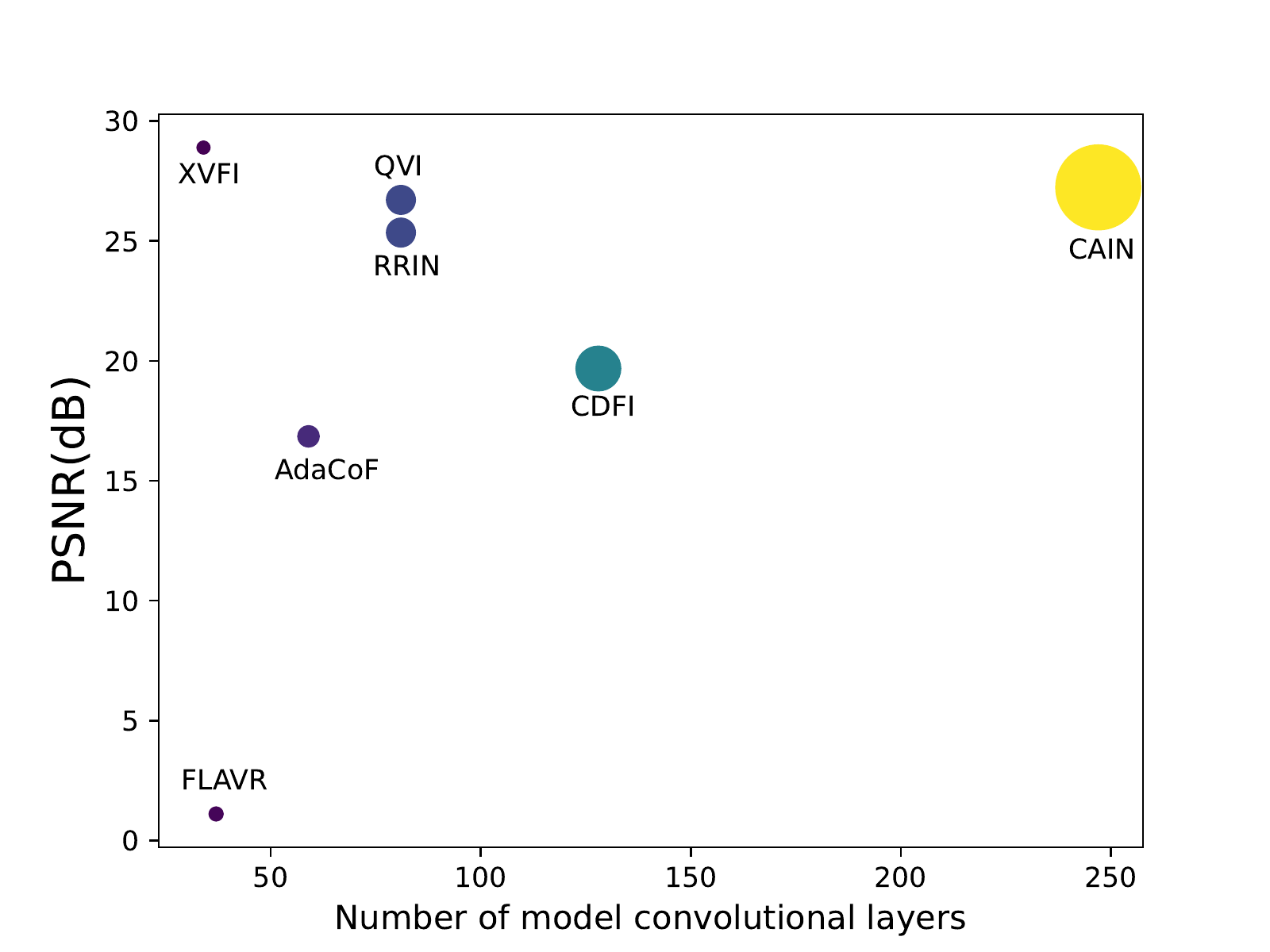}}
		\end{minipage}
	\end{center}
	\caption{Comparison of the number of model parameters and convolutional layers of video frame interpolation models of different PSNR of $\alpha=0.02$ on UCF-101 dataset.}
	\label{fig:model_size}
\end{figure*}

\begin{figure}[t]
	\begin{center}
		\centering
		\includegraphics[width=0.8\linewidth]{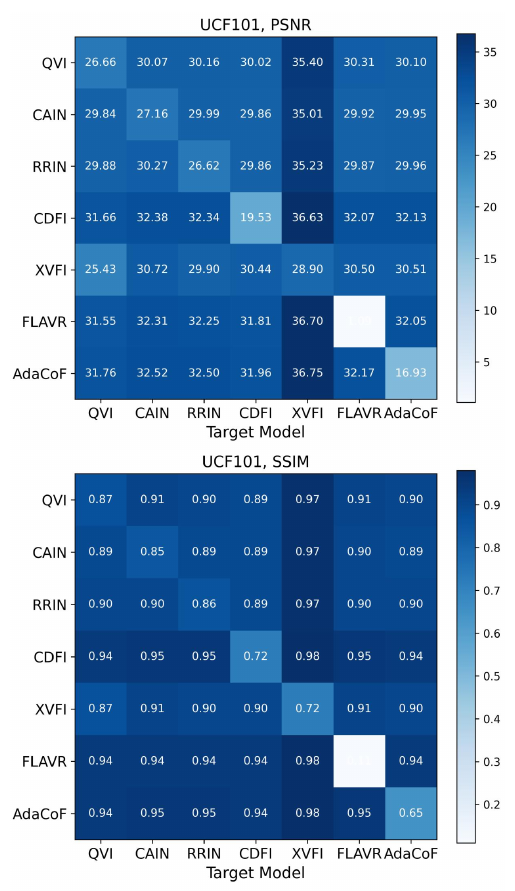}
	\end{center}
	\caption{Comparison of the transferbility of IAA adversarial examples across models in terms of PSNR and SSIM when $\alpha=0.02$ on UCF-101 dataset.}
	\label{fig:transfer}
\end{figure}

\subsection{Attack Performance}

We evaluate the performance of basic attack and our proposed attack algorithm with UCF-101, Vimeo90K and Middlebury benchmark. Figure \ref{fig:psnr_and_ssim_comparison} shows the IAA attack performance on UCF-101 dataset in terms of PSNR and SSIM with different $\alpha$ values. PSNR and SSIM both decreases rapidly as $\alpha$ increases on all video interpolation models which is consistent with our intuition. For example, in terms of AdaCoF model, the PSNR/SSIM values for $\alpha$ = 0.01 and 0.08 are 20.14/0.76 and 13.61/0.45 with our IAA method, respectively. Figure \ref{fig:visual_comparison_alpha} shows the visualized results of attacked intermediate frames as $\alpha$ increasing. It is noticeable that among all VIF models, FLAVR is specially vulnerable to adversarial examples. This may be contributed to its 3D space-time convolutions which contain the least convolutional layers. The statistics reveal that the VIF models are vulnerable to adversarial attacks.

When it comes to the comparison of different attack methods, Figure \ref{fig:psnr_and_ssim_comparison} also shows that our proposed method exhibits comparable performance to basic attack while decreasing the attack cost. When $\alpha$ is $0.08$, the quality degradation even becomes much more severe with our method which reduces the PSNR/SSIM values of most tested models except XVFI more than basic attack on UCF-101 dataset. This result is attributed to the exploitation of similarity between consecutive frames which means the adversarial perturbation inherits properties from previous frames. Figure \ref{fig:visual_all_models} shows the visualized results of IAA attack on different video frame interpolation models. Though $\alpha$ is only $0.02$, the generated intermediate frames are wrapped so severe that it is impossible to be used in the video quality enhancement task.

\textbf{Transferability.} Figure \ref{fig:transfer} summarizes the transferability for deep learning-based video interpolation models on UCF-101 benchmark when $\alpha$ = 0.02. Transferability represents the possibility that a misclassified adversarial example is also misclassified by another classifier in image classification task. In this paper, we also evaluate the transferability of video interpolation task by using adversarial examples from other source models as inputs for the target model. And we measure the PSNR and SSIM value of the output intermediate frames in the experiment. The figure \ref{fig:transfer} shows that the adversarial examples do not transfer very well between different models. But we can see that the adversarial examples generated by QVI, XVFI, CAIN and RRIN can transfer better than the rest methods. One possible reason for the poor transferability is that lower level vision tasks like video frame interpolation generalize poorly on its original task. The inherit property determines the poor transferability on the task. However, the outputs' PSNR values are still been declined by the perturbation generated by other source models to some extent. 

\textbf{Relation to model size.} It can be seen that the adversarial robustness of different video frame interpolation models is slightly related to their model sizes. Figure \ref{fig:model_size} shows the relationship between model size and model's robustness in terms of the number of model parameters and convolutional layers. For example, CDFI, a compressed version based on AdaCoF, performs better than its larger version in terms of PSNR values on adversarial examples. However, for models with different network architectures, the model size and robustness of deep learning-based video interpolation models is almost irrelevant.

\begin{figure}[t]
	\begin{center}
		\centering
		\includegraphics[width=1.0\linewidth]{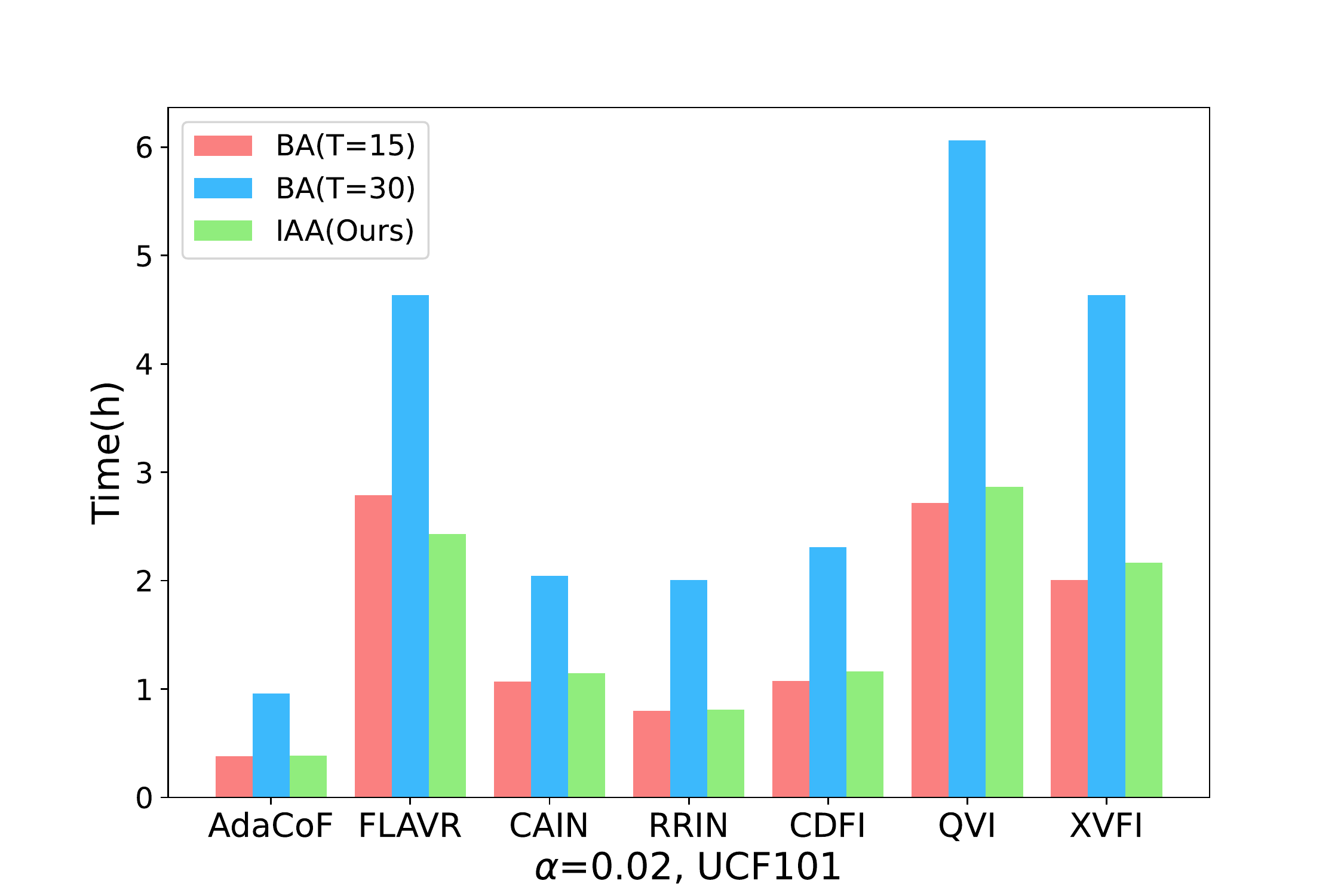}
	\end{center}
	\caption{Comparison of the attack time in terms of UCF-101 dataset when $\alpha=0.02$. }
	\label{fig:time}
\end{figure}

\textbf{Attack time.} The major goal of our IAA method is to improve attack efficiency for generating adversarial examples for tons of frames in the video for frame interpolation tasks. We record the time cost to obtain high-performance adversarial examples on UCF-101 dataset. Figure \ref{fig:time} shows that our method only spends almost half of the time to complete the attack process while achieving the same degradation on PSNR/SSIM values which means that it has become much easier for us to attack a whole video in practice.

\begin{table}[tb]
\centering
  \caption{Ablation study of our proposed attack. We compare the PSNR and SSIM values of target models under $BA_{random}$ and IAA when $\alpha=0.02$ on UCF-101.}
  \label{tab:ablation}
  \small
  \setlength{\tabcolsep}{0.6em} 
  \begin{tabular}{c c c c}
    \toprule
    Target Model & Methods &  PSNR$\Downarrow$  & SSIM$\Downarrow$ \\
    \hline
    \centering \multirow{2}{*}{QVI \cite{QVI}} 
    & $BA_{random}$ & 27.56 & 0.8705 \\ 
    & $IAA(Ours)$ & $\bm{26.71}$ & $\bm{0.8155}$ \\
    \hline
    \centering\multirow{2}{*}{CAIN \cite{CAIN}}
    & $BA_{random}$ & 29.46 & $\bm{0.8041}$ \\ 
    & $IAA(Ours)$ & $\bm{27.25}$ & 0.8522 \\
    \hline
    \centering \multirow{2}{*}{RRIN \cite{RRIN}} 
    & $BA_{random}$ & 27.17 & $\bm{0.8138}$ \\ 
    & $IAA(Ours)$ & $\bm{25.35}$ & 0.8596 \\
    \hline
    \centering \multirow{2}{*}{XVFI \cite{XVFI}} 
    & $BA_{random}$ & 31.50 & $\bm{0.5224}$ \\ 
    & $IAA(Ours)$ & $\bm{28.90}$ & 0.7202 \\
    \hline
    \centering \multirow{2}{*}{CDFI \cite{CDFI}} 
    & $BA_{random}$ & 31.32 & 0.8748 \\ 
    & $IAA(Ours)$ & $\bm{19.81}$ & $\bm{0.7281}$ \\
    \hline
    \centering \multirow{2}{*}{AdaCoF \cite{AdaCoF}} 
    & $BA_{random}$ & 29.55 & 0.8407 \\ 
    & $IAA(Ours)$ & $\bm{16.92}$ & $\bm{0.6532}$ \\
    \hline
    \centering \multirow{2}{*}{FLAVR \cite{FLAVR}} 
    & $BA_{random}$ & 16.95 & 0.4795 \\ 
    & $IAA(Ours)$ & $\bm{1.09}$ & $\bm{0.1078}$ \\
    \bottomrule
  \end{tabular}
\vspace{-2mm}
\end{table}

\subsection{Ablation Study}
Experiments above has shown the effectiveness of IAA on VIF models, it achieves the best attack performance on most models in terms of PSNR, SSIM and attack time. We then do the ablation study to investigate the role of our proposed improved part playing in the attack. Based on basic attack, we initialize perturbation pairs for each intermediate frame as $P_0 \sim \mathbb{N}(0, 1)$ and constrain it in $[-\alpha, \alpha]$. Noted that, to evaluate the contribution of IAA, the attack iteration for basic attack here is the same as IAA which is 15. \\
\indent\setlength{\parindent}{2em} Table \ref{tab:ablation} shows the results of the attack with $\alpha=0.02$ on UCF-101. Although each initialized perturbation in $BA_{random}$ are given a value which is comparable to IAA, the attack performance is not been improved evidently. For example, on CDFI model which is a compressed version of AdaCoF, the PSNR and SSIM is 31.32 and 0.8748 which means that the generated intermediate frames are still of high quality after $BA_{random}$ attack. But under the same settings, the metrics deteriorate to 19.81 and 0.7281 at the same time with IAA. In general, our proposed IAA performs much better than $BA_{random}$ which shows that it is our proposed initialization method improve the attack performance.

\section{Advanced topic}

\begin{figure*}[t]
	\begin{center}
		\centering
		\begin{minipage}[b]{0.13\linewidth}
			\centering
			\centerline{\includegraphics[width=1.0\linewidth]{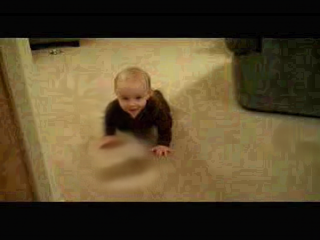}}
		\end{minipage}
  		\begin{minipage}[b]{0.13\linewidth}
			\centering
			\centerline{\includegraphics[width=1.0\linewidth]{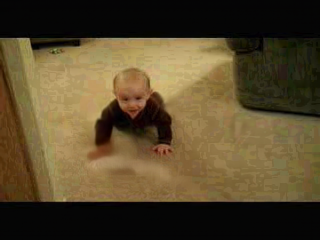}}
		\end{minipage}
		\begin{minipage}[b]{0.13\linewidth}
			\centering
			\centerline{\includegraphics[width=1.0\linewidth]{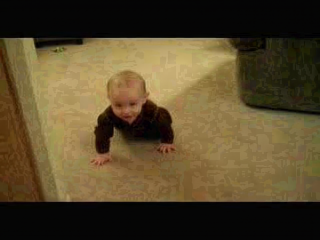}}
		\end{minipage}
  		\begin{minipage}[b]{0.13\linewidth}
			\centering
			\centerline{\includegraphics[width=1.0\linewidth]{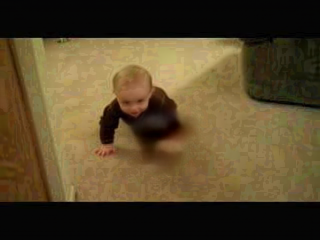}}
		\end{minipage}
  		\begin{minipage}[b]{0.13\linewidth}
			\centering
			\centerline{\includegraphics[width=1.0\linewidth]{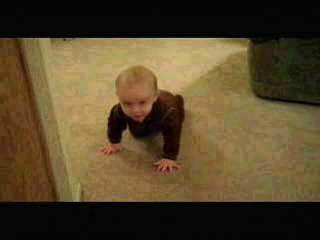}}
		\end{minipage}
  		\begin{minipage}[b]{0.13\linewidth}
			\centering
			\centerline{\includegraphics[width=1.0\linewidth]{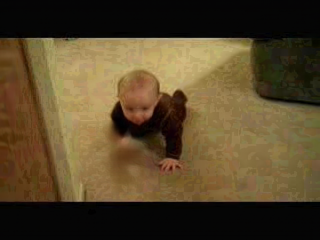}}
		\end{minipage}
		\begin{minipage}[b]{0.13\linewidth}
			\centering
			\centerline{\includegraphics[width=1.0\linewidth]{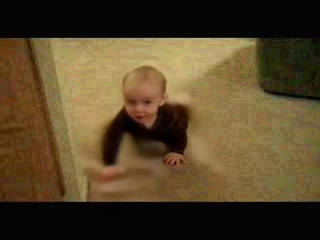}}
		\end{minipage}
  		\begin{minipage}[b]{0.13\linewidth}
			\centering
			\centerline{\includegraphics[width=1.0\linewidth]{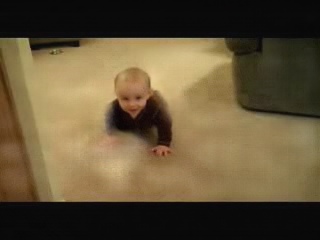}}
		\end{minipage}
    		\begin{minipage}[b]{0.13\linewidth}
			\centering
			\centerline{\includegraphics[width=1.0\linewidth]{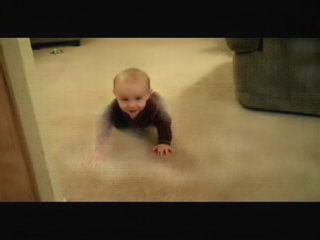}}
		\end{minipage}
		\begin{minipage}[b]{0.13\linewidth}
			\centering
			\centerline{\includegraphics[width=1.0\linewidth]{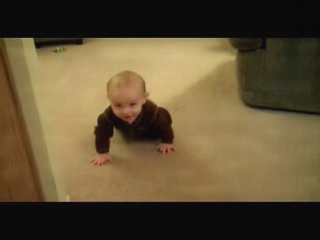}}
		\end{minipage}
    		\begin{minipage}[b]{0.13\linewidth}
			\centering
			\centerline{\includegraphics[width=1.0\linewidth]{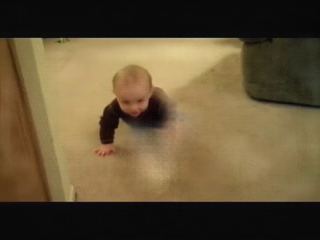}}
		\end{minipage}
  		\begin{minipage}[b]{0.13\linewidth}
			\centering
			\centerline{\includegraphics[width=1.0\linewidth]{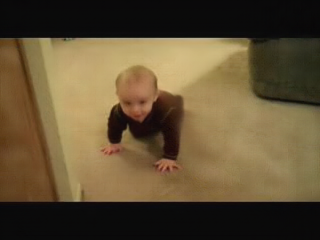}}
		\end{minipage}
    	\begin{minipage}[b]{0.13\linewidth}
			\centering
			\centerline{\includegraphics[width=1.0\linewidth]{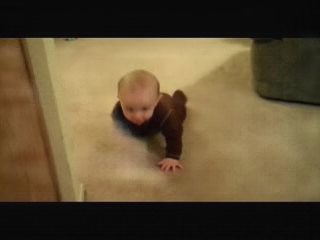}}
		\end{minipage}
		\begin{minipage}[b]{0.13\linewidth}
			\centering
			\centerline{\includegraphics[width=1.0\linewidth]{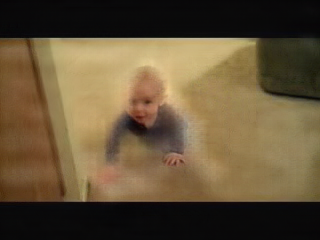}}
		\end{minipage}
    		\begin{minipage}[b]{0.13\linewidth}
			\centering
			\centerline{\includegraphics[width=1.0\linewidth]{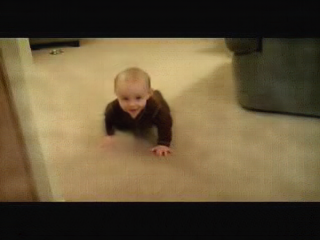}}
		\end{minipage}
    		\begin{minipage}[b]{0.13\linewidth}
			\centering
			\centerline{\includegraphics[width=1.0\linewidth]{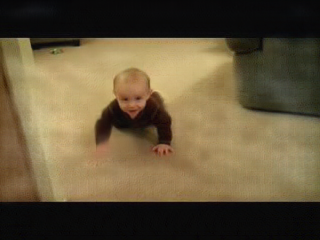}}
		\end{minipage}
		\begin{minipage}[b]{0.13\linewidth}
			\centering
			\centerline{\includegraphics[width=1.0\linewidth]{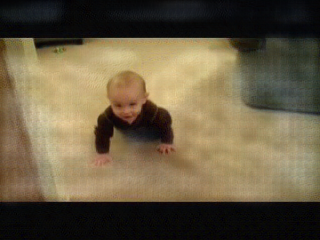}}
		\end{minipage}
    		\begin{minipage}[b]{0.13\linewidth}
			\centering
			\centerline{\includegraphics[width=1.0\linewidth]{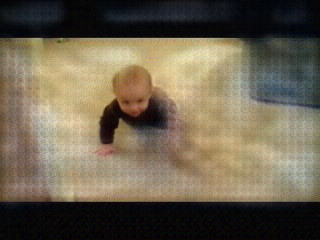}}
		\end{minipage}
  		\begin{minipage}[b]{0.13\linewidth}
			\centering
			\centerline{\includegraphics[width=1.0\linewidth]{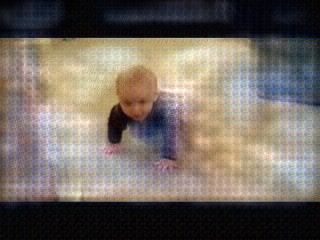}}
		\end{minipage}
    	\begin{minipage}[b]{0.13\linewidth}
			\centering
			\centerline{\includegraphics[width=1.0\linewidth]{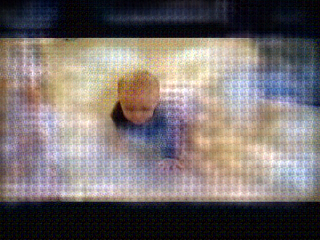}}
		\end{minipage}
		\begin{minipage}[b]{0.13\linewidth}
			\centering
			\centerline{\includegraphics[width=1.0\linewidth]{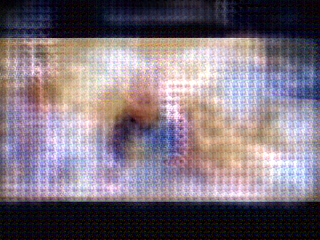}}
		\end{minipage}
	\end{center}
	\caption{Visualization of targeted attack on UCF-101 when $\alpha=0.02$. We select seven consecutive generated intermediate frames from one video. The first row is the results of non-targeted attack. The second and third row represents the targeted attack based on basic attack and IAA, respectively.}
	\label{fig:visual_targetd_attack}
\end{figure*}

\subsection{Targeted Attack}
In our experiments, we apply white image as the target image because such target images with extreme values usually make people uncomfortable. Figure \ref{fig:visual_targetd_attack} shows the visualization results of the targeted attack. It is obvious that the visually attack performance for targeted attack is better than non-targeted attack. And it can be observed that the first two attack performance is relatively similar visually compared to the IAA. When it comes to IAA based attack, the visual attack performance become much better and The latter frames in the video is destroyed much more serious than the former ones visually. The same results are not observed on the other two attacks. The main reason is that for BA-based targeted attack, the attack iteration is not enough. But IAA-based targeted attack successfully inherit the information from previous ones which make up for the drawbacks of attack iteration. \\
\indent\setlength{\parindent}{2em} Table \ref{tab:target_tab1} shows the comparison of PSNR and SSIM values of VIF models under targeted attacks on UCF-101. When $\alpha=0.04$, IAA performs the best on most VIF models except XVFI. The PSNR and SSIM results for targeted attack is similar to non-targeted attacks in which IAA achieves better attack performance while recuding the attack cost.

\begin{table}[tb]
\centering
  \caption{Comparison of the targeted attack performance on VIF models with $\alpha=0.04$ on UCF-101.}
  \label{tab:target_tab1}
  \small
  \setlength{\tabcolsep}{0.6em} 
  \begin{tabular}{c rr rr rr}
    \toprule
    \multirow{2}{*}{Target Model} &\multicolumn{2}{c}{BA(T=15)} & \multicolumn{2}{c}{BA(T=30)} & \multicolumn{2}{c}{IAA} \\
    \cmidrule{2-7}
    & PSNR$\Downarrow$  & SSIM$\Downarrow$ & PSNR$\Downarrow$  & SSIM$\Downarrow$ & PSNR$\Downarrow$  & SSIM$\Downarrow$ \\
    \midrule
    \centering 
    QVI \cite{QVI} & 27.95 & 0.8976 & 21.85 & 0.8631 & $\bm{17.98}$ & $\bm{0.6325}$ \\ 
    \hline
    \centering
    CAIN \cite{CAIN} & 26.91 & 0.8373 & 23.95 & 0.8003 & $\bm{17.66}$ & $\bm{0.5930}$ \\ 
    \hline
    \centering 
    RRIN \cite{RRIN}& 26.82 & 0.8493 & 23.50 & 0.8033 & $\bm{19.14}$ & $\bm{0.7371}$ \\ 
    \hline
    \centering 
    XVFI \cite{XVFI} & 29.08 & 0.3185 & $\bm{27.05}$ & $\bm{0.2851}$ & 28.27 & 0.3131 \\ 
    \hline
    \centering 
    CDFI \cite{CDFI} & 32.31 & 0.9105 & 31.20 & 0.8818 & $\bm{27.03}$ & $\bm{0.7646}$ \\ 
    \hline
    \centering 
    AdaCoF \cite{AdaCoF} & 29.93 & 0.8914 & 25.96 & 0.8320 & $\bm{19.86}$ & $\bm{0.6736}$ \\ 
    \hline
    \centering
    FLAVR \cite{FLAVR} & 18.46 & 0.6375 & 12.50 & 0.4154 & $\bm{8.40}$ & $\bm{0.2680}$ \\
    \bottomrule
  \end{tabular}
\vspace{-2mm}
\end{table}

\begin{table}[tb]
\centering
  \caption{Comparison of the attack performance on video recognition models with $\alpha=0.04$.}
  \label{tab:recogni_table1}
  \small
  \setlength{\tabcolsep}{0.6em} 
  \begin{tabular}{c c c c}
    \toprule
   Methods & Target Model & $Acc(\%)\Downarrow$  & $Time(s)\Downarrow$ \\
    \midrule
    \centering
    No attack & C3D & 78.67 & 3781 \\
    \hline
    \centering 
    BA(T=15) & C3D & 27.27 & $\bm{30755}$ \\
    \hline
    \centering 
    BA(T=30) & C3D & 20.53 & 56923 \\
    \hline
    \centering 
    IAA & C3D & $\bm{7.48}$ & 33131 \\
    \bottomrule
  \end{tabular}
\vspace{-2mm}
\end{table}

\subsection{Attack Transferability to Video Recognition Models}
We adopt C3D as the target model to attack in our experiments. The model is trained on sports-1m dataset and fine-tuned on UCF-101 training set. We evaluate the accuracy and attack time of basic attack (BA) and IAA $\alpha \in \{0.01, 0.02, 0.04, 0.08\}$ and $eps \in \{0.0005, 0.001, 0.002, 0.003\}$ on UCF-101 test set. We adopt the iteration number as $T \in \{15, 30\}$. \\

\begin{table}[htbp]\caption{Comparison of IAA attack performance on C3D with different $\alpha$ on UCF-101.} \centering
\begin{tabular}{c c c c c c }
\hline
$\alpha$ & 0 & 0.01 & 0.02 & 0.04 & 0.08   \\
\hline
$Acc(\%)\Downarrow$ & 78.67 & 52.02  & 24.82  & 7.48  & 4.52  \\
\hline
$Time(s)\Downarrow$  & 3792 & 33157  & 33501   & 33131   & 33212 \\
\hline
\end{tabular}\label{tab:recogni_table2}
\end{table}

\indent\setlength{\parindent}{2em} Table \ref{tab:recogni_table2} reveals the results of our proposed method with different $\alpha$ on C3D. With the increase of $\alpha$, the attack performance improves. The classification accuracy of C3D deteriorates to $7.48\%$ when $\alpha=0.04$ and $4.52$ with $\alpha=0.08$ which means that our attack has disabled the classifier successfully. When it comes to the comparsion of basic attack and IAA method, the experiments shows IAA still performs better than basic attack. Table \ref{tab:recogni_table1} shows the details of the attack performance in terms of different attack methods. Contributed to the inherited information from former frames, IAA's attack performance achieves $13.05\%$ better than BA of $T = 30$ while spending the same time as BA of $T = 15$. In summary, the experiments show that IAA transfers very well to video recognition models and it can successfully fool video recognition models while cost much less computation.

\section{Conclusion}
We first customize Projected Gradient Descent (PGD) method, a widely used adversarial attack, to deep learning-based video interpolation (VIF) models. Then we investigate the adversarial robustness of deep learning-based VIF models with different model properties. For improving attack efficiency, we propose a novel attack named Inter-frame Accelerate Attack (IAA) to accelerate the attack process on VIF models by making full use of the similarity between consecutive frames in the video. Our experiments show that VIF models are vulnerable to adversarial attacks and our proposed method IAA achieves better performance than basic attack while saving considerable computation resources. We show that targeted attack performs better visual quality degradation on VIF models. Furthermore, we show the great transferability of our proposed attack to higher level vision task such as video recognition. In other words, though our method is simple, it shows excellent performance on VIF models and great tranferability to the video recognition model.


\bibliographystyle{ACM-Reference-Format}
\bibliography{referenece}

\end{document}